\documentclass[10pt,twocolumn,letterpaper]{article}

\usepackage[pagenumbers]{cvpr} 

\usepackage{cuted} 
\usepackage{multirow}
\usepackage{currfile} 

\usepackage{caption} 

\usepackage{amsmath}
\usepackage{amssymb}
\usepackage{enumitem}

\definecolor{cvprblue}{rgb}{0.21,0.49,0.74}
\usepackage[pagebackref,breaklinks,colorlinks,allcolors=cvprblue]{hyperref}

\def\confName{CVPR}
\def\confYear{2027}

\title{HandQuery: Dense Vertex Queries for Multi-View 4D Hand Reconstruction}

\author{
Yi Gu\textsuperscript{\dag1}
\and
Xuanhua Chen\textsuperscript{\dag1}
\and
Yupei Zeng\textsuperscript{1}
\and
Hanyang Cao\textsuperscript{1}
\and
Mingxi Xu\textsuperscript{1}%
\and
Yukang Gao\textsuperscript{1}
\and
Yangchen Zhou\textsuperscript{1}
\and
Fanghai Zhang\textsuperscript{1}
\and
Weiming Zhang\textsuperscript{1}
\and
Jiaxu Wang\textsuperscript{1}
\and
Kang Du\textsuperscript{1}
\and
Zhengyang Shen\textsuperscript{2}
\and
Renjing Xu\textsuperscript{1}
\and
\\
\textsuperscript{1}HKUST(GZ) \quad \textsuperscript{2} Google \quad \textsuperscript{\dag}Equal Contributions
}

\begin{document}
\maketitle
\setlength{\stripsep}{4pt}  
\begin{strip}
\centering
\includegraphics[width=\linewidth]{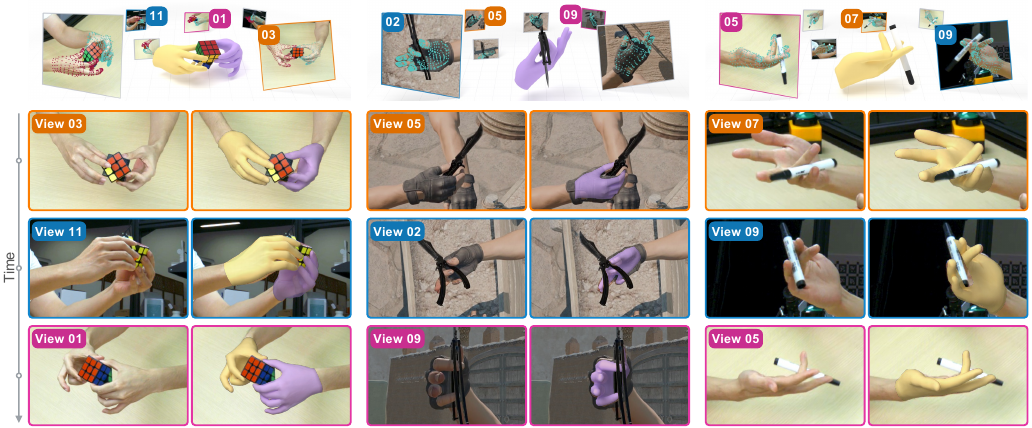}
\captionsetup{hypcap=false}
\captionof{figure}{\textbf{Zero-shot 4D hand reconstruction across real and virtual environments.} We showcase three unseen multi-view sequences: two real-world studio captures (left and right) and one from \textit{Counter-Strike 2} (middle). The top row illustrates the multi-view observations, while the bottom rows show input images alongside our reconstructions at different viewpoints and time steps. HandQuery accurately recovers articulated hand motion across diverse camera configurations, appearances, and interactions without additional training.}
\captionsetup{hypcap=true}

\label{fig:\currfilebase}%
\end{strip}
\begin{abstract}
High-accuracy hand motion capture typically relies on marker-based optical systems or instrumented gloves, yet applications such as visual imitation learning and hand-object interaction analysis require faithful visual observations of bare hands. Existing markerless methods often process views independently or aggregate them frame by frame, leaving complementary evidence across cameras and time underexploited. We present \textbf{HandQuery}, a dense 4D hand reconstruction framework for synchronized and calibrated multi-view videos. HandQuery integrates spatial, temporal, and geometry-aware cross-view fusion within the visual encoder, then uses independent MANO vertex queries to predict dense 2D surface landmarks, visibility, and uncertainty. This formulation supports flexible vertex selection during both training and inference. Uncertainty-aware multi-view MANO fitting integrates the predicted 2D observations into 4D hand meshes. To provide consistent dense supervision and broaden training diversity, we construct synthetic multi-view hand videos with varied appearances and camera configurations. Experiments demonstrate strong accuracy across five real-world benchmarks and robustness to fewer cameras. In zero-shot evaluation on a newly captured studio dataset, HandQuery reduces mean per-vertex error by approximately 39\% relative to the strongest baseline. Code and models: \href{https://reconworldlab.github.io/handquery/}{\textcolor{magenta}{project page}}. 
\end{abstract}    
\vspace{-8pt}               
\section{Introduction}
\label{sec:introduction}

High-accuracy hand motion capture typically relies on marker-based
optical systems or instrumented gloves. However, applications such as visual imitation learning, hand-object interaction, and photorealistic digital-human modeling require both accurate motion and faithful visual observations of bare hands. Markers and gloves alter the appearance of the hands and can obscure contact regions or interfere with natural manipulation. This motivates markerless hand capture that preserves the visual evidence of natural interactions while recovering detailed hand motion.

Recent monocular estimators~\cite{pavlakos2024reconstructing,potamias2025wilor,khirodkar2026sapiens2} recover hand meshes from individual images, but their application to multi-camera capture typically relies on independent per-image predictions followed by triangulation or model fitting. Dedicated multi-view methods~\cite{zhang2021direct,yang2023poem,yang2024mlphand,yang2025multi,kuang2026jump} aggregate evidence across cameras, yet many still operate frame by frame. These approaches leave complementary information across views and time underexploited during prediction. This limitation is particularly consequential for hands, whose small image footprint, fine articulation, and frequent self-occlusion and interaction-induced occlusion make individual observations inherently incomplete.

Recently, dense landmark-based methods~\cite{cuevas2026mamma,wood20223d,2024lookma,patel2025camerahmr} have demonstrated impressive performance in markerless motion capture. However, joint prediction of dense hand landmarks across views and time remains comparatively underexplored. Representative pipelines such as MAMMA~\cite{cuevas2026mamma} and LookMa~\cite{2024lookma} estimate landmarks from individual images, incorporating cross-view information during subsequent reconstruction. Under severe occlusion, post-hoc fitting may be unable to correct ambiguous or systematically biased observations. This motivates fusing multi-view temporal evidence before estimating dense hand landmarks.

We introduce \textbf{HandQuery}, a framework for dense 4D hand
reconstruction from synchronized and calibrated multi-view videos. HandQuery integrates spatial, temporal, and geometry-aware cross-view fusion within the visual encoder, allowing observations from different cameras and neighboring frames to inform landmark prediction. A vertex that is occluded or ambiguous in one image can thus be localized using complementary spatiotemporal evidence. We decode the fused features using independent vertex queries. Each query uses a learned identity tied only to a MANO~\cite{MANO:SIGGRAPHASIA:2017} vertex and shared across views and frames. The decoder predicts the corresponding 2D location, visibility, and uncertainty in each view and frame without interactions between vertex queries. Consequently, training can sample random vertex subsets, while inference can request all 778 MANO vertices or any desired subset. Finally, uncertainty-aware multi-view MANO fitting integrates the predicted observations into a 4D hand mesh.

Reliable supervision is a second challenge. Accurately annotated multi-view hand videos are scarce, and misaligned reference meshes can compromise dense supervision from projected vertices. Motivated by the generalization benefits of synthetic hand data~\cite{si2026anyhand,shen2026vephand,li2023renderih}, we re-render hand-object and inter-hand motions under diverse viewpoints, backgrounds, and lighting conditions. The resulting corpus thus supports reliable surface supervision while broadening visual diversity to improve generalization.

HandQuery achieves state-of-the-art performance on five real-world benchmarks, outperforming both strong monocular estimators followed by multi-view fitting and dedicated multi-view reconstruction methods. To evaluate generalization beyond these benchmarks, we collect HandQuery-Real and curate HandQuery-Eval, a manually verified subset for quantitative zero-shot evaluation. Without dataset-specific fine-tuning, HandQuery reduces MPVPE by 44–47\% over WiLoR + Fit and approximately 39\% over MAMMA trained with the same data and fitter. Zero-shot qualitative comparisons on SHOW3D~\cite{rim2026show3d} and GigaHands~\cite{fu2025gigahands} further reveal cases where our reconstructions align more closely with the images than the provided annotations. 

Our contributions are:
\begin{itemize}
    \item We present a 4D hand reconstruction framework that integrates spatial, temporal, and geometry-aware cross-view fusion with independent vertex queries, enabling flexible vertex selection during both training and inference.

    \item We construct a synthetic multi-view hand video corpus with spatiotemporally consistent dense annotations and collect a real-world multi-view studio dataset for evaluating zero-shot generalization.

    \item We demonstrate strong reconstruction accuracy across five real-world benchmarks, robustness to fewer cameras, and substantial improvements over existing methods in zero-shot evaluation on unseen capture setups.
\end{itemize}


\section{Related Work}
\label{sec:related_work}
\noindent
\textbf{Monocular Hand Reconstruction.} Monocular hand reconstruction has progressed from image-based estimators such as HaMeR~\cite{pavlakos2024reconstructing} and WiLoR~\cite{potamias2025wilor} to video-based approaches. HandFlow~\cite{xu2026handflow} and StableHand~\cite{zeng2026stablehand} employ flow matching for temporally coherent hand motion estimation, while recent works~\cite{wang2026surprising,wu2026perceive,moon2026rlhnd,liu2026ace} leverage video diffusion models for robust 4D hand reconstruction. When applied to multi-view videos, monocular estimators typically process views independently and combine predictions through triangulation or model fitting. Such post-hoc fusion can struggle to correct unreliable predictions under severe occlusions and visual ambiguities, motivating joint reasoning across views and time before prediction.

\noindent
\textbf{Multi-view Hand Reconstruction.}
Multi-view methods leverage complementary observations and geometric constraints to improve hand reconstruction. POEM~\cite{yang2023poem,yang2025multi} and MLPHand~\cite{yang2024mlphand} aggregate geometric features across views, while JUMP-Hand~\cite{kuang2026jump} and UA-Fit~\cite{zakour2026learn} explicitly model observation uncertainty for robust reconstruction. HGGT~\cite{liu2026hggt} further enables joint hand geometry and camera pose estimation from uncalibrated images. UST-Hand~\cite{han2026usthand} models spatiotemporal relationships through probabilistic point clouds, while HOST~\cite{peng2026high} integrates multi-view temporal cues for hand-object reconstruction. Closely related to our work, MAMMA~\cite{cuevas2026mamma} predicts dense 2D surface landmarks from individual images, followed by cross-view matching and SMPL-X fitting. Unlike its image-wise landmark estimation, HandQuery integrates multi-view spatiotemporal evidence before predicting dense landmarks.

\noindent
\textbf{Synthetic Hand Data.}
Obtaining accurate ground-truth annotations for dynamic scenes is challenging, motivating the widespread use of synthetic data for supervising 4D vision tasks~\cite{greff2022kubric,zheng2023pointodyssey,karaev2023dynamicstereo,zhao2026synthverse}. For hand reconstruction, existing synthetic datasets~\cite{li2023renderih,zimmermann2017learning,hasson2019learning,si2026anyhand,shen2026vephand} primarily focus on image-level supervision, offering diverse hand poses and appearances but limited support for joint multi-view temporal learning. Some efforts have extended synthetic hand data to dynamic and multi-view settings. Re:InterHand~\cite{moon2023dataset} provides relighted multi-view sequences of interacting hands, while MAMMA~\cite{cuevas2026mamma} introduces MammaSyn, primarily designed for full-body motion capture. However, jointly achieving diverse camera configurations and synchronized multi-view temporal observations remains underexplored for hand-centric reconstruction. To this end, we introduce \textbf{HandQuery-Syn}, which re-renders hand-object and hand-hand motions under diverse camera viewpoints, backgrounds, and lighting conditions while preserving consistent dense annotations across views and time.

\section{Method}
\label{sec:method}

\begin{figure*}[!t] 
\centering
\includegraphics[width=1.0\linewidth]{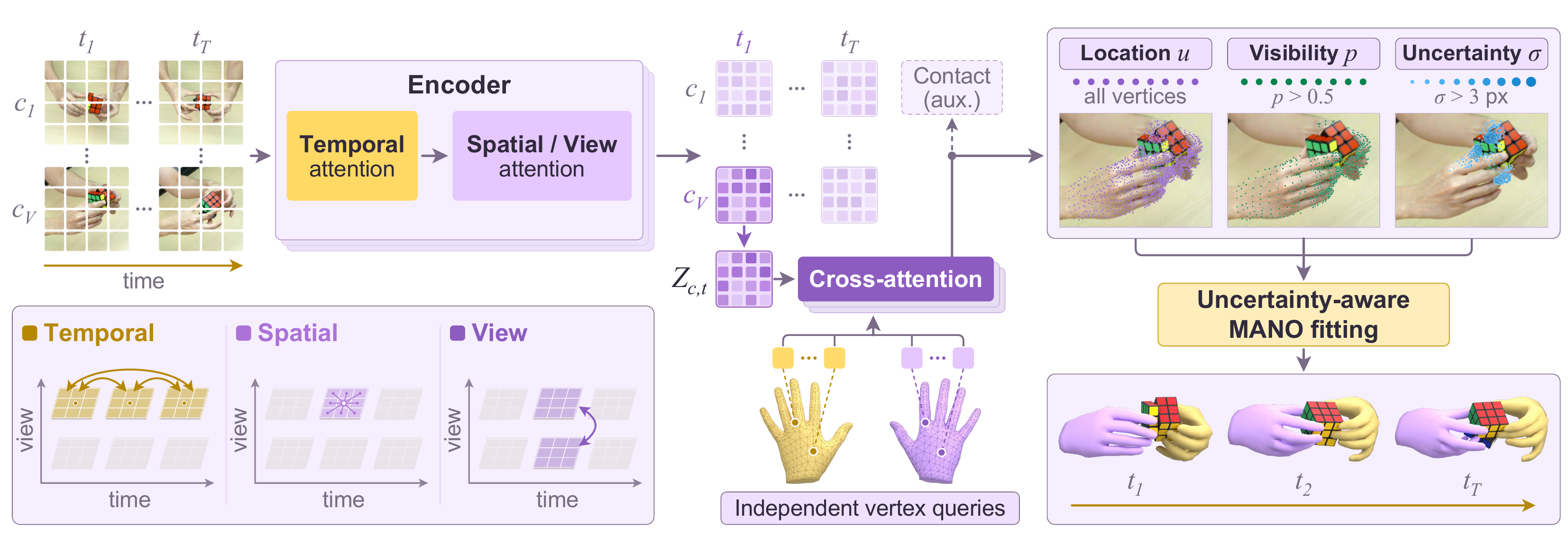}
\caption{\textbf{Overview of HandQuery.} Given synchronized and calibrated multi-view videos, our encoder aggregates complementary evidence across views and time through temporal, spatial, and view attention (bottom left). Independent MANO vertex queries then cross-attend to the contextualized features, predicting dense 2D landmark locations, visibility, and uncertainty, with contact prediction as auxiliary supervision. Finally, uncertainty-aware MANO fitting integrates these observations to recover 4D hand motion.}
\label{fig:pipeline} 
\end{figure*}

Our goal is to recover 4D hand geometry from synchronized and calibrated multi-view videos. The key formulation is to use dense 2D surface landmarks as the intermediate representation: each landmark corresponds to the projection of one of the $778$ MANO vertices, providing explicit image-space evidence for subsequent multi-view 4D hand reconstruction. A common landmark-based paradigm, exemplified by MAMMA~\cite{cuevas2026mamma}, predicts each view and frame independently and aggregates the resulting measurements only afterward. Such late fusion compresses an image into point estimates before other observations can help interpret it: an occluded or locally ambiguous vertex in one view cannot be disambiguated using a clearer view or a nearby frame during feature extraction.  Our key idea is therefore to exchange complementary evidence before landmark prediction.  We first build a multi-view video encoder in which features communicate across synchronized views and along time (Sec.~\ref{subsection:encoder}). We then use independent MANO-vertex queries to read the fused representation and predict a vertex-indexed set of 2D landmark predictions (Sec.~\ref{subsection:decoder}). The per-view predictions retain explicit image evidence and known vertex correspondence, allowing calibrated multi-view geometry and MANO fitting to recover the final mesh. Finally, to improve generalization to unseen capture domains and camera layouts, we construct a synthetic multi-view video dataset for training (Sec.~\ref{subsection:handquery_sync}).

\subsection{Multi-view Spatiotemporal Encoding}
\label{subsection:encoder}

\textbf{Pre-processing.}
For each view and frame, we extract an $S\times S$ perspective hand crop
using the supplied bounding box.  The crop is the image of a virtual camera
whose optical axis follows the box center, with a focal length fixed within
each view's clip. We retain the corresponding intrinsics $K^{\mathrm{crop}}_{c,t}$ and extrinsics $E^{\mathrm{crop}}_{c,t}$ to keep image transformations consistent with camera geometry. Even for a static physical rig, hand motion therefore changes the virtual camera orientations, exposing the encoder to motion-dependent projective configurations. The resulting crops center the hand throughout the clip while preserving calibrated relations across views.

\noindent
\textbf{Spatiotemporal feature exchange.} To aggregate complementary evidence across views and time, we encode the crops with a shared pretrained ViT~\cite{dosovitskiy2021an, xu2022vitpose} using spatial and temporal positional embeddings. We follow the terminology of MV-TAP~\cite{koo2026mv}, referring to the three operations as temporal, spatial, and view attention, while applying them to image-patch tokens to construct contextual features for dense landmark prediction. Inspired by the encoder design of Depth Anything 3 (DA3)~\cite{lin2026depth,lin2025depth}, we enable within-image and cross-view processing by regrouping tokens within the existing ViT blocks, with the same pretrained attention weights shared across both scopes. Specifically, \emph{spatial attention} exchanges information among patches within a single view and frame, relating local appearance to the surrounding hand context. \emph{View attention} jointly processes all patch tokens from synchronized views at a fixed time, allowing ambiguous image regions to draw evidence from other viewpoints. To incorporate temporal context efficiently, we separate temporal aggregation from spatial and cross-view feature exchange, following the divided attention principle of TimeSformer~\cite{gberta_2021_ICML}. In each encoder block, \emph{temporal attention} first aggregates features across frames at a fixed view and patch index. The updated features then undergo either spatial or view attention, allowing temporal context to inform subsequent within-image or cross-view interactions. We organize the encoder into an initial stage of within-view spatiotemporal encoding and a subsequent stage of multi-view fusion. The former uses spatial attention, while the latter alternates between view and spatial attention. We set the stage depths to $L_{\mathrm{intra}}:L_{\mathrm{multi}}=1:2$, with $L=L_{\mathrm{intra}}+L_{\mathrm{multi}}$.

For geometric conditioning, we apply PRoPE~\cite{li2026cameras} only within view attention. PRoPE encodes relative camera projections, making this geometric conditioning invariant to a global change of world coordinate frame. We compute the encoding from the updated crop-camera intrinsics and extrinsics to maintain consistency with hand motion and geometric augmentation.

\noindent
\textbf{Preserving observation structure.} We organize the encoder output with explicit view, frame, and patch dimensions:
\begin{equation}
  \mathbf{Z} = \mathcal{E}\!\left(
    \mathbf{I}^{\mathrm{crop}};\mathcal{G}^{\mathrm{crop}}\right)
  \in\mathbb{R}^{V\times T\times P\times D},
  \label{eq:structured_features}
\end{equation}
where $\mathbf{I}^{\mathrm{crop}}$ denotes the collection of hand crops from $V$ synchronized views over $T$ frames,
$\mathcal{G}^{\mathrm{crop}}$ denotes their normalized camera geometry, $P$ is the number of patches per crop, and $D$ is the feature dimension. Each feature $\mathbf{z}_{c,t,p}$ incorporates multi-view temporal context while remaining associated with patch $p$ in view $c$ at frame $t$. The corresponding features $\mathbf{Z}_{c,t}$ serve as the memory for the independent landmark queries in Sec.~\ref{subsection:decoder}, specifying the target image in which vertices are to be localized.

\subsection{Independent Dense Landmark Queries}
\label{subsection:decoder}
MAMMA~\cite{cuevas2026mamma} predicts a predefined set of surface landmarks, with $63$ per hand, and models their dependencies through decoder self-attention. Our qualitative comparison of landmark densities suggests that denser coverage benefits challenging hand configurations, even when improvements in average reconstruction error are modest. These observations motivate a denser set of surface constraints for highly articulated hands, whose local configurations may be insufficiently constrained by a small landmark set. We therefore define queries for all 778 MANO vertices, providing dense correspondence across the hand surface.

Scaling query self-attention to this dense vocabulary, however, introduces a cost quadratic in the number of queries. Following D4RT~\cite{zhang2026efficiently}, we instead decode each vertex independently through cross-attention to its target image features.  Since the encoder has already integrated evidence across views and time, the decoder can read each landmark from a contextualized representation. A learned embedding $\mathbf{s}_i$ identifies the vertex, while the selected memory
$\mathbf{Z}_{c,t}=\{\mathbf{z}_{c,t,p}\}_{p=1}^{P}$ identifies the target
camera $c$ and frame $t$:
\begin{equation}
  \mathbf{h}_{i,c,t}
  = \mathcal{D}\!\left(\mathbf{s}_i,\mathbf{Z}_{c,t}\right).
  \label{eq:independent_readout}
\end{equation}
The vertex embedding is shared across cameras and frames.  The decoder
comprises six blocks of cross-attention and pointwise feed-forward layers,
all reading the same image memory. Our decoder comparisons show that independent queries achieve reconstruction accuracy comparable to coupled decoding. For fixed encoder features, each prediction is also invariant to the other queries being evaluated.  This allows us to sample only $64$ queries per image during training and decode all $778$ vertices of the target hand at inference. Selected vertices can likewise be evaluated in separate batches to accommodate available memory.

Linear heads parameterize the landmark measurement
\begin{equation}
  \hat{\mathbf{y}}_{i,c,t}
  = \left(\hat{\mathbf{u}}_{i,c,t},\hat{\sigma}_{i,c,t},
    \hat{p}_{i,c,t},\hat{\mathbf{p}}^{\mathrm{contact}}_{i,c,t}\right),
  \label{eq:heads}
\end{equation}
where $\hat{\mathbf{u}}$, $\hat{\sigma}$, $\hat{p}$, and
$\hat{\mathbf{p}}^{\mathrm{contact}}$ denote 2D location, uncertainty,
visibility, and hand-object/hand-hand contact, respectively. The location, uncertainty, and visibility drive multi-view model fitting, and the contact heads provide auxiliary interaction supervision.

\subsection{Synthetic Multi-view Video Data}
\label{subsection:handquery_sync}
Reliable supervision for multi-view 4D hand reconstruction requires images, camera geometry, and hand annotations to describe the same evolving scene. In real capture pipelines, small calibration and synchronization errors can compound with inaccuracies in intermediate estimates and annotation, potentially introducing inconsistencies between the reference annotations and observed hand motion (See Supp. Sec 2). Such discrepancies introduce conflicting supervision across views and time, undermining the consistency that reconstruction is intended to exploit. Camera coverage presents a separate limitation: even large video collections can be tied to a small set of capture rigs, exposing the model to a restricted distribution of viewpoints and relative camera geometries. These limitations motivate synthetic videos that pair consistent geometric supervision with diverse camera configurations.

We replay MANO sequences and available object trajectories from the training splits of DexYCB, HO3D, OakInk, ARCTIC, and InterHand2.6M~\cite{chao2021dexycb,hampali2020honnotate,yang2022oakink,fan2023arctic,moon2020interhand2}, excluding evaluation sequences. We attach procedural forearms~\cite{si2026anyhand} with optional sleeves to the hand meshes, providing visual context and occlusions around the wrist. Each clip is rendered with randomized skin appearance~\cite{gao2022dart}, clothing, scene materials, and lighting under a newly sampled set of calibrated cameras. Images and labels are generated from the same scene geometry, providing projections of all $778$ MANO vertices, visibility, and hand-object/hand-hand contact labels. More details are available in the supplementary.

Synthetic rendering still leaves an appearance gap to real captures. We therefore mix real and synthetic clips during training, combining real image statistics with the consistent geometric supervision and broader camera coverage provided by synthetic data.
\section{Experiments}
\label{sec:experiments}

\subsection{Datasets and Metrics}
\label{sec:datasets_and_metrics}
Following POEM-v2~\cite{yang2025multi}, we train on five multi-view hand datasets: DexYCB~\cite{chao2021dexycb}, HO3D~\cite{hampali2020honnotate}, OakInk~\cite{yang2022oakink},
ARCTIC~\cite{fan2023arctic}, and InterHand2.6M~\cite{moon2020interhand2}, and evaluate on the same held-out sequences. We report mean per-vertex position error (MPVPE) over all 778 MANO vertices in calibrated rig coordinates, together with its wrist-aligned (RR-V) and Procrustes-aligned (PA-V) variants. AUC-V measures the area under the vertex PCK curve in absolute coordinates, using 20 uniformly spaced thresholds in [0, 20] mm ([0, 50] mm for HO3D-MV), following POEM~\cite{yang2023poem}. We additionally report the corresponding joint errors (MPJPE*, RR-J*, and PA-J*) using POEM's 21-joint definition, comprising MANO-regressed joints and five fingertip vertices. All distance errors are measured in millimeters and averaged over evaluated hand-frames. \textbf{MPVPE} and \textbf{PA-V} are our primary metrics, independent of joint definitions. See Sec. 1 of the supplementary material for implementation details.


\begin{figure}[t]
\centering
{\scriptsize%
\makebox[\dimexpr\columnwidth/6\relax]{Image}%
\makebox[\dimexpr\columnwidth/6\relax]{GT}%
\makebox[\dimexpr\columnwidth/6\relax]{WiLoR + Fit}%
\makebox[\dimexpr\columnwidth/6\relax]{POEM-v2}%
\makebox[\dimexpr\columnwidth/6\relax]{UA-Fit}%
\makebox[\dimexpr\columnwidth/6\relax]{Ours$^*$}}\par\vspace{2pt}

\includegraphics[width=\columnwidth]{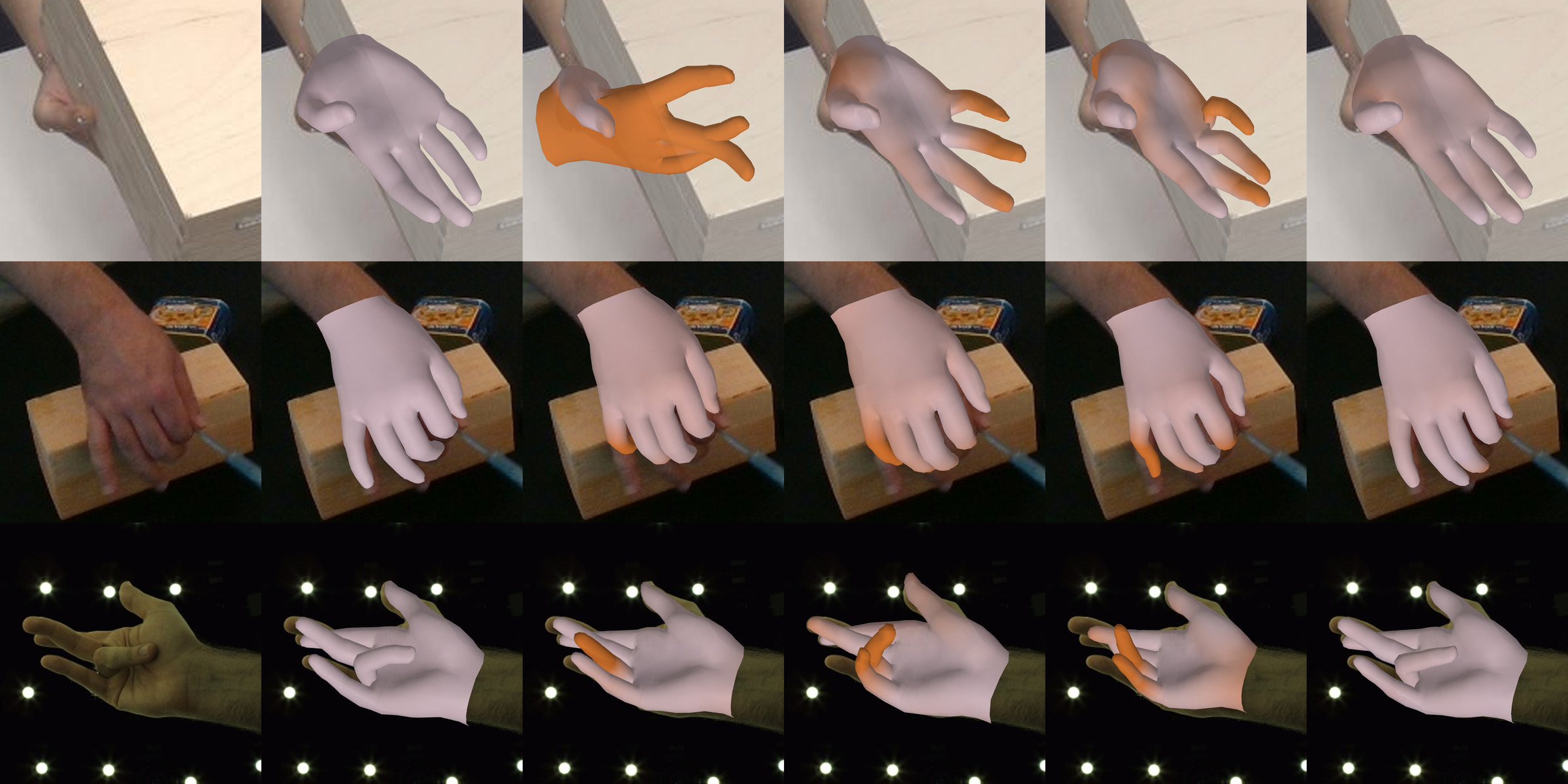}

\caption{Qualitative comparison on ARCTIC-MV, DexYCB-MV,
and InterHand-MV (rows 1--3). Predicted meshes are overlaid on the input images and colored by their 3D per-vertex errors relative to GT, from the base hand color (0\,mm) to orange ($\geq 20$\,mm).}

\label{fig:qual_err}
\end{figure}

\subsection{Comparison with Prior Methods}
\label{sec:comparisons}

\begin{table*}[tp]
\centering
\small
\definecolor{hqRankFirst}{HTML}{FF8080}
\definecolor{hqRankSecond}{HTML}{FFBF80}
\definecolor{hqRankThird}{HTML}{FFF79A}
\setlength{\tabcolsep}{0pt}
\setlength{\fboxsep}{0pt}
\renewcommand{\arraystretch}{1}
\newcommand{\hqRank}[2]{\colorbox{#1}{\makebox[.10\textwidth]{\strut#2}}}
\newcommand{\hqKey}[2]{\colorbox{#1}{\kern.25em\strut#2\kern.25em}}
\setlength{\abovecaptionskip}{4pt}

\caption[Five multi-view benchmarks]{
Five multi-view benchmarks (mm; \hqKey{hqRankFirst}{first}, \hqKey{hqRankSecond}{second}, and \hqKey{hqRankThird}{third}). Ours: five real datasets; Ours$^*$: additionally synthetic clips (Sec.~\ref{subsection:handquery_sync}). $^\S$: MAMMA~\cite{cuevas2026mamma}, retrained for hands on Ours$^*$'s data using our fitter. $^\dagger$: evaluated with released checkpoints; others: published (--: unreported). Joints$^*$: POEM definition. $^\P$: POEM-v2's released InterHand crop boxes can be incorrectly enlarged by out-of-image joints. We fix their center and scale using in-image joints (19.2\% of views; MPVPE 7.62 $\to$ 5.92\,mm).
}

\begin{tabular}{ll*{7}{c}}
\toprule
\makebox[.14\textwidth][l]{Dataset (frames)} & \makebox[.16\textwidth][l]{Method} & MPVPE$\downarrow$ & RR-V$\downarrow$ & PA-V$\downarrow$ & AUC-V$\uparrow$ & MPJPE$^*\downarrow$ & RR-J$^*\downarrow$ & PA-J$^*\downarrow$ \\
\midrule
\multirow{11}{*}{\shortstack[l]{DexYCB-MV\\(4,951)}} & MvP & 9.77 & 12.18 & 8.14 & 0.530 & 6.23 & 9.47 & 4.26 \\
 & POEM & 6.13 & 7.21 & 4.00 & 0.700 & 6.06 & 7.30 & 3.93 \\
 & MLPHand & 6.20 & 7.30 & 4.15 & -- & 6.23 & 7.41 & 4.11 \\
 & WiLoR + Fit\smash{$^\dagger$} & 7.58 & 7.41 & 3.85 & 0.629 & 7.57 & 7.55 & 3.82 \\
 & POEM-v2\smash{$^\dagger$} & 6.44 & 7.50 & 3.81 & 0.684 & 6.26 & 7.82 & 3.66 \\
 & Sapiens2 + Fit\smash{$^\dagger$} & 8.81 & 9.81 & 6.18 & 0.578 & 8.65 & 9.81 & 5.97 \\
 & JUMP-Hand & \hqRank{hqRankThird}{5.45} & \hqRank{hqRankThird}{6.41} & 3.77 & \hqRank{hqRankThird}{0.730} & \hqRank{hqRankThird}{5.31} & \hqRank{hqRankThird}{6.41} & 3.61 \\
 & UA-Fit\smash{$^\dagger$} & 7.86 & 8.91 & 4.92 & 0.617 & 7.55 & 8.61 & 4.60 \\
 & MAMMA\smash{$^\S$} & 6.84 & 11.97 & \hqRank{hqRankThird}{3.72} & 0.663 & 6.41 & 12.39 & \hqRank{hqRankThird}{3.55} \\
\cmidrule(l){2-9}
 & Ours & \hqRank{hqRankSecond}{4.56} & \hqRank{hqRankSecond}{5.31} & \hqRank{hqRankSecond}{2.91} & \hqRank{hqRankSecond}{0.775} & \hqRank{hqRankSecond}{4.44} & \hqRank{hqRankSecond}{5.35} & \hqRank{hqRankSecond}{2.85} \\
 & Ours$^*$ & \hqRank{hqRankFirst}{4.38} & \hqRank{hqRankFirst}{5.09} & \hqRank{hqRankFirst}{2.77} & \hqRank{hqRankFirst}{0.784} & \hqRank{hqRankFirst}{4.26} & \hqRank{hqRankFirst}{5.12} & \hqRank{hqRankFirst}{2.72} \\
\midrule
\multirow{10}{*}{\shortstack[l]{HO3D-MV\\(2,706)}} & MvP & 20.95 & 27.08 & 10.04 & 0.590 & 18.72 & 24.90 & 10.44 \\
 & POEM & 17.20 & 21.45 & 9.97 & 0.660 & 17.28 & 21.94 & 9.60 \\
 & MLPHand & 18.69 & 23.28 & 10.54 & -- & 18.70 & 23.76 & 10.11 \\
 & POEM-v2\smash{$^\dagger$} & 7.41 & 10.13 & 4.53 & 0.852 & 6.93 & 10.24 & 4.08 \\
 & Sapiens2 + Fit\smash{$^\dagger$} & 9.83 & 11.64 & 6.90 & 0.804 & 9.04 & 11.52 & 6.13 \\
 & JUMP-Hand & 13.39 & 16.96 & 8.78 & 0.730 & 13.10 & 17.07 & 8.30 \\
 & UA-Fit\smash{$^\dagger$} & 7.81 & \hqRank{hqRankThird}{8.74} & 5.10 & 0.844 & 7.22 & \hqRank{hqRankThird}{8.36} & 4.33 \\
 & MAMMA\smash{$^\S$} & \hqRank{hqRankThird}{6.77} & 10.72 & \hqRank{hqRankThird}{4.01} & \hqRank{hqRankThird}{0.865} & \hqRank{hqRankThird}{6.30} & 10.77 & \hqRank{hqRankThird}{3.52} \\
\cmidrule(l){2-9}
 & Ours & \hqRank{hqRankSecond}{6.11} & \hqRank{hqRankSecond}{7.13} & \hqRank{hqRankSecond}{3.79} & \hqRank{hqRankSecond}{0.878} & \hqRank{hqRankSecond}{5.81} & \hqRank{hqRankSecond}{6.81} & \hqRank{hqRankSecond}{3.31} \\
 & Ours$^*$ & \hqRank{hqRankFirst}{5.63} & \hqRank{hqRankFirst}{6.81} & \hqRank{hqRankFirst}{3.45} & \hqRank{hqRankFirst}{0.887} & \hqRank{hqRankFirst}{5.30} & \hqRank{hqRankFirst}{6.52} & \hqRank{hqRankFirst}{2.96} \\
\midrule
\multirow{9}{*}{\shortstack[l]{OakInk-MV\\(4,978)}} & MvP & 9.69 & 11.75 & 7.74 & 0.530 & 7.32 & 9.99 & 4.97 \\
 & POEM & 6.20 & 7.63 & 4.21 & 0.700 & 6.01 & 7.46 & 4.00 \\
 & MLPHand & 6.17 & 7.49 & 4.26 & -- & 6.00 & 7.52 & 4.09 \\
 & WiLoR + Fit\smash{$^\dagger$} & 9.19 & 10.19 & 5.41 & 0.548 & 8.84 & 9.93 & 4.83 \\
 & Sapiens2 + Fit\smash{$^\dagger$} & 10.87 & 13.39 & 6.48 & 0.530 & 10.27 & 13.08 & 5.71 \\
 & JUMP-Hand & \hqRank{hqRankThird}{5.94} & \hqRank{hqRankThird}{6.99} & \hqRank{hqRankThird}{4.13} & \hqRank{hqRankThird}{0.720} & \hqRank{hqRankThird}{5.72} & \hqRank{hqRankThird}{6.94} & \hqRank{hqRankThird}{3.86} \\
 & MAMMA\smash{$^\S$} & 7.69 & 12.99 & 4.23 & 0.618 & 7.19 & 13.47 & 3.90 \\
\cmidrule(l){2-9}
 & Ours & \hqRank{hqRankSecond}{5.47} & \hqRank{hqRankSecond}{6.14} & \hqRank{hqRankSecond}{3.49} & \hqRank{hqRankSecond}{0.727} & \hqRank{hqRankSecond}{5.18} & \hqRank{hqRankSecond}{5.99} & \hqRank{hqRankSecond}{3.14} \\
 & Ours$^*$ & \hqRank{hqRankFirst}{5.26} & \hqRank{hqRankFirst}{5.86} & \hqRank{hqRankFirst}{3.37} & \hqRank{hqRankFirst}{0.737} & \hqRank{hqRankFirst}{4.97} & \hqRank{hqRankFirst}{5.67} & \hqRank{hqRankFirst}{3.00} \\
\midrule
\multirow{7}{*}{\shortstack[l]{ARCTIC-MV\\(17,393)}} & WiLoR + Fit\smash{$^\dagger$} & 7.55 & 12.69 & 5.74 & 0.641 & 6.74 & 12.82 & 5.08 \\
 & POEM-v2\smash{$^\dagger$} & 5.80 & \hqRank{hqRankThird}{7.13} & 4.38 & 0.713 & 5.34 & \hqRank{hqRankThird}{6.83} & 3.87 \\
 & Sapiens2 + Fit\smash{$^\dagger$} & 10.39 & 16.54 & 7.09 & 0.517 & 9.26 & 17.05 & 6.38 \\
 & UA-Fit\smash{$^\dagger$} & 7.54 & 9.52 & 5.66 & 0.629 & 6.62 & 8.89 & 4.77 \\
 & MAMMA\smash{$^\S$} & \hqRank{hqRankThird}{5.53} & 9.08 & \hqRank{hqRankThird}{4.02} & \hqRank{hqRankThird}{0.724} & \hqRank{hqRankThird}{4.89} & 9.07 & \hqRank{hqRankThird}{3.46} \\
\cmidrule(l){2-9}
 & Ours & \hqRank{hqRankSecond}{4.85} & \hqRank{hqRankSecond}{7.03} & \hqRank{hqRankSecond}{3.78} & \hqRank{hqRankSecond}{0.758} & \hqRank{hqRankSecond}{4.26} & \hqRank{hqRankSecond}{6.74} & \hqRank{hqRankSecond}{3.22} \\
 & Ours$^*$ & \hqRank{hqRankFirst}{4.48} & \hqRank{hqRankFirst}{6.38} & \hqRank{hqRankFirst}{3.41} & \hqRank{hqRankFirst}{0.776} & \hqRank{hqRankFirst}{3.94} & \hqRank{hqRankFirst}{6.09} & \hqRank{hqRankFirst}{2.89} \\
\midrule
\multirow{7}{*}{\shortstack[l]{InterHand-MV\\(85,255)}} & WiLoR + Fit\smash{$^\dagger$} & 4.83 & \hqRank{hqRankThird}{6.69} & 3.24 & 0.760 & 4.43 & 6.93 & 2.86 \\
 & POEM-v2\smash{$^{\dagger,\P}$} & 5.92 & 7.13 & 4.71 & 0.716 & 5.50 & 6.91 & 4.25 \\
 & Sapiens2 + Fit\smash{$^\dagger$} & 5.63 & 6.85 & 4.61 & 0.721 & 4.86 & \hqRank{hqRankThird}{6.43} & 3.65 \\
 & UA-Fit\smash{$^\dagger$} & 8.77 & 14.92 & 6.23 & 0.577 & 7.79 & 15.13 & 5.44 \\
 & MAMMA\smash{$^\S$} & \hqRank{hqRankThird}{4.60} & 6.87 & \hqRank{hqRankThird}{3.07} & \hqRank{hqRankThird}{0.773} & \hqRank{hqRankThird}{4.28} & 6.52 & \hqRank{hqRankThird}{2.62} \\
\cmidrule(l){2-9}
 & Ours & \hqRank{hqRankSecond}{3.96} & \hqRank{hqRankSecond}{4.74} & \hqRank{hqRankSecond}{2.72} & \hqRank{hqRankSecond}{0.803} & \hqRank{hqRankSecond}{3.61} & \hqRank{hqRankSecond}{4.70} & \hqRank{hqRankSecond}{2.29} \\
 & Ours$^*$ & \hqRank{hqRankFirst}{3.68} & \hqRank{hqRankFirst}{4.46} & \hqRank{hqRankFirst}{2.59} & \hqRank{hqRankFirst}{0.817} & \hqRank{hqRankFirst}{3.29} & \hqRank{hqRankFirst}{4.43} & \hqRank{hqRankFirst}{2.09} \\
\bottomrule
\end{tabular}
\vspace{-2pt}

\label{tab:exp_main}
\end{table*}


\noindent\textbf{Baselines.}
We compare against three groups of methods. The first group includes multi-view methods trained \textbf{separately on each benchmark dataset}: MvP~\cite{zhang2021direct}, POEM~\cite{yang2023poem},
MLPHand~\cite{yang2024mlphand}, and JUMP-Hand~\cite{kuang2026jump}. We report their published results and reproduce MvP and POEM on OakInk-MV using released checkpoints. The second group consists of POEM-v2 (POEM-v2 Large)~\cite{yang2025multi} and UA-Fit~\cite{zakour2026learn}, whose released models are trained on \textbf{mixtures containing all five benchmark datasets}. POEM-v2 and UA-Fit are omitted on OakInk-MV due to different evaluation protocols. The third group comprises image-wise estimators, including WiLoR~\cite{potamias2025wilor},
Sapiens2~\cite{khirodkar2026sapiens2}, and
MAMMA~\cite{cuevas2026mamma}. For a stronger comparison, we retrain MAMMA with 256 landmarks per hand on the same real and synthetic training data as Ours*, and evaluate it using our fitter, which outperforms MAMMA's original fitter across all five datasets. WiLoR is excluded from HO3D-MV because its training data overlap the test sequences.

\noindent\textbf{Results.}
As shown in Table~\ref{tab:exp_main}, Ours achieves the lowest MPVPE across all five benchmarks even without synthetic training, while Ours* further reduces the error by 3.8--7.9\%. These results suggest that multi-view temporal fusion and synthetic supervision provide complementary benefits. The comparison with MAMMA is particularly informative: despite being trained with the same real and synthetic data as Ours* and using our fitter, MAMMA still has 10--50\% higher MPVPE than Ours trained on real data alone. This suggests that stronger supervision and downstream fitting alone are insufficient to close the gap, highlighting the benefit of exchanging multi-view temporal evidence before landmark prediction. Fig.~\ref{fig:qual_err} further demonstrates our robustness to object and self-occlusions, especially for partially occluded fingers. Model size and inference cost comparisons are provided in Sec.~4.1 of the supplementary material.


\begin{table}[t]
\centering
\small
\setlength{\tabcolsep}{0pt}
\setlength{\abovecaptionskip}{4pt}
\setlength{\aboverulesep}{0pt}
\setlength{\belowrulesep}{0pt}
\renewcommand{\arraystretch}{1.22}

\caption{MPVPE (mm) with fewer cameras, on the hand-frames of Table~\ref{tab:exp_main}. All methods use the same fixed camera subsets.}

\begin{tabular*}{\columnwidth}{@{\hspace{1pt}}c@{\hspace{3pt}\vrule width \arrayrulewidth\hspace{5pt}}l@{\hspace{7pt}\vrule width \arrayrulewidth\extracolsep{\fill}}rr@{\extracolsep{0pt}\hspace{2pt}}l@{\extracolsep{\fill}}r@{\extracolsep{0pt}\hspace{2pt}}l@{\extracolsep{\fill}}r@{\extracolsep{0pt}\hspace{2pt}}l@{\extracolsep{\fill}}@{}}
\toprule
\multicolumn{1}{@{\hspace{1pt}}c@{\hspace{3pt}\hspace{\arrayrulewidth}\hspace{5pt}}}{} & Cameras & \multicolumn{1}{c}{8} & \multicolumn{2}{c}{4} & \multicolumn{2}{c}{3} & \multicolumn{2}{c}{2} \\
\midrule
\multirow{4}{*}{\rotatebox[origin=c]{90}{DexYCB}} & WiLoR + Fit & 7.58 & 9.03 & {\scriptsize\textcolor{black!55}{+19\%}} & 8.70 & {\scriptsize\textbf{+15\%}} & 13.07 & {\scriptsize\textcolor{black!55}{+72\%}} \\
 & POEM-v2 & 6.44 & 7.38 & {\scriptsize\textcolor{black!55}{+15\%}} & 7.73 & {\scriptsize\textcolor{black!55}{+20\%}} & 10.44 & {\scriptsize\textcolor{black!55}{+62\%}} \\
 & UA-Fit & 7.86 & 8.87 & {\scriptsize\textcolor{black!55}{+13\%}} & 9.13 & {\scriptsize\textcolor{black!55}{+16\%}} & 13.41 & {\scriptsize\textcolor{black!55}{+71\%}} \\
 & Ours$^*$ & \textbf{4.38} & \textbf{4.83} & {\scriptsize\textbf{+10\%}} & \textbf{5.11} & {\scriptsize\textcolor{black!55}{+17\%}} & \textbf{6.59} & {\scriptsize\textbf{+51\%}} \\
\midrule
\multirow{4}{*}{\rotatebox[origin=c]{90}{ARCTIC}} & WiLoR + Fit & 7.55 & 8.89 & {\scriptsize\textcolor{black!55}{+18\%}} & 9.94 & {\scriptsize\textcolor{black!55}{+32\%}} & 12.70 & {\scriptsize\textcolor{black!55}{+68\%}} \\
 & POEM-v2 & 5.80 & 6.42 & {\scriptsize\textcolor{black!55}{+11\%}} & 6.88 & {\scriptsize\textcolor{black!55}{+19\%}} & 9.08 & {\scriptsize\textcolor{black!55}{+56\%}} \\
 & UA-Fit & 7.54 & 8.58 & {\scriptsize\textcolor{black!55}{+14\%}} & 9.23 & {\scriptsize\textcolor{black!55}{+22\%}} & 12.80 & {\scriptsize\textcolor{black!55}{+70\%}} \\
 & Ours$^*$ & \textbf{4.48} & \textbf{4.85} & {\scriptsize\textbf{+8\%}} & \textbf{5.23} & {\scriptsize\textbf{+17\%}} & \textbf{6.23} & {\scriptsize\textbf{+39\%}} \\
\bottomrule
\end{tabular*}

\label{tab:exp_cams}
\end{table}

\subsection{Number of Cameras}
\label{sec:cams}
We evaluate robustness to reduced camera coverage on DexYCB-MV and ARCTIC-MV, both recorded with eight cameras. We evaluate all methods using the same fixed subsets of
four, three, and two cameras without retraining
(Table~\ref{tab:exp_cams}). As the number of cameras decreases from eight to two, our MPVPE increases by 51\% and 39\% on DexYCB-MV and ARCTIC-MV, respectively, compared with 56--72\% for the baselines. Notably, with only three cameras, our method still outperforms all evaluated baselines using eight cameras (5.11 vs. 6.44 mm on DexYCB-MV and 5.23 vs. 5.80 mm on ARCTIC-MV). This robustness suggests that leveraging temporal context
can help compensate for missing viewpoints, allowing
reliable reconstruction even with sparse camera coverage.

\subsection{Zero-Shot Evaluation}
\label{sec:zeroshot}

\begin{table}[t]
\centering
\footnotesize
\setlength{\tabcolsep}{0pt}
\setlength{\abovecaptionskip}{4pt}
\setlength{\aboverulesep}{0pt}
\setlength{\belowrulesep}{0pt}
\renewcommand{\arraystretch}{1.22}


\caption{Zero-shot comparison on HandQuery-Eval with 8 and 4 of the 12 cameras (best per camera count in bold).}

\begin{tabular*}{\columnwidth}{@{\hspace{1pt}}c@{\hspace{3pt}\vrule width \arrayrulewidth\hspace{4pt}}l@{\hspace{5pt}\vrule width \arrayrulewidth\extracolsep{\fill}}cccc@{}c@{\hspace{2pt}\vrule width \arrayrulewidth\hspace{2pt}}ccc@{\hspace{1pt}}}
\toprule
\multicolumn{1}{@{\hspace{1pt}}c@{\hspace{3pt}\hspace{\arrayrulewidth}\hspace{4pt}}}{} & & \multicolumn{4}{c}{Vertices} & & \multicolumn{3}{c@{\hspace{1pt}}}{Joints$^*$} \\
\cmidrule{3-6}\cmidrule{8-10}
\multicolumn{1}{@{\hspace{1pt}}c@{\hspace{3pt}\hspace{\arrayrulewidth}\hspace{4pt}}}{} & Method & MPVPE & RR & PA & AUC$\uparrow$ & & MPJPE & RR & \multicolumn{1}{c@{\hspace{1pt}}}{PA} \\
\midrule
\multirow{6}{*}{\rotatebox[origin=c]{90}{8 cameras}} & POEM-v2 & 9.91 & 12.42 & 5.57 & 0.594 &  & 9.50 & 12.58 & 5.22 \\
 & UA-Fit & 9.49 & 10.17 & 5.63 & 0.585 &  & 9.07 & 10.10 & 5.04 \\
 & WiLoR + Fit & 5.21 & 5.80 & 3.24 & 0.748 &  & 5.06 & 5.84 & 3.01 \\
 & MAMMA & 4.80 & 7.60 & 3.06 & 0.765 &  & 4.48 & 7.87 & 2.88 \\
\cmidrule{2-10}
 & Ours & 7.42 & 8.62 & 3.71 & 0.704 &  & 7.05 & 8.71 & 3.49 \\
 & Ours$^*$ & \textbf{2.93} & \textbf{3.72} & \textbf{2.32} & \textbf{0.856} &  & \textbf{2.74} & \textbf{3.67} & \textbf{2.11} \\
\midrule
\multirow{6}{*}{\rotatebox[origin=c]{90}{4 cameras}} & POEM-v2 & 11.02 & 13.65 & 6.22 & 0.558 &  & 10.51 & 13.69 & 5.81 \\
 & UA-Fit & 10.50 & 11.50 & 5.99 & 0.560 &  & 10.01 & 11.45 & 5.38 \\
 & WiLoR + Fit & 5.76 & 6.54 & 3.62 & 0.728 &  & 5.65 & 6.62 & 3.41 \\
 & MAMMA & 5.02 & 7.96 & 3.18 & 0.754 &  & 4.69 & 8.24 & 2.99 \\
\cmidrule{2-10}
 & Ours & 8.10 & 9.47 & 3.99 & 0.696 &  & 7.70 & 9.60 & 3.78 \\
 & Ours$^*$ & \textbf{3.04} & \textbf{3.78} & \textbf{2.39} & \textbf{0.850} &  & \textbf{2.85} & \textbf{3.72} & \textbf{2.18} \\
\bottomrule
\end{tabular*}

\label{tab:exp_real}
\end{table}

We evaluate generalization to capture setups that are not seen in training, quantitatively on our HandQuery-Eval and qualitatively on SHOW3D~\cite{rim2026show3d} and GigaHands~\cite{fu2025gigahands}.

\noindent\textbf{HandQuery-Eval.} We collect HandQuery-Real, a multi-view dataset of hand-object interaction recorded with 12 synchronized and calibrated cameras at $1624\times1240$ pixels, comprising over 2,000 segments and approximately 650k frames. All sequences are annotated using a semi-automatic, agent-assisted pipeline followed by manual verification. We select 39 challenging segments (16,572 frames) with frequent occlusions and fast motions, spanning five camera layouts, to form HandQuery-Eval. Further details on data collection and annotation are provided in Sec.~3 of the supplementary material.

\noindent\textbf{Quantitative results.}
Table~\ref{tab:exp_real} reveals several insights into zero-shot generalization. Without synthetic augmentation, Ours outperforms POEM-v2 and UA-Fit but trails WiLoR + Fit, highlighting the value of diverse in-the-wild visual priors. Incorporating our synthetic data substantially improves generalization, reducing MPVPE over Ours by 60.5\% and 62.5\% with eight and four cameras, respectively. Ours* further outperforms WiLoR + Fit by 44\% and 47\%, achieving the best performance across all metrics. These gains suggest that diverse synthetic multi-view training helps bridge the domain gap between existing benchmarks and unseen capture setups. Notably, compared with MAMMA trained on the same real and synthetic data and evaluated using our fitter, Ours* achieves approximately 39\% lower MPVPE in both settings, demonstrating the benefits of our architecture beyond synthetic supervision and fitting.

\begin{figure}[t]
\centering
{\scriptsize
\makebox[\dimexpr\columnwidth/7\relax]{Image}%
\makebox[\dimexpr\columnwidth/7\relax]{Reference}%
\makebox[\dimexpr\columnwidth/7\relax]{WiLoR+Fit}%
\makebox[\dimexpr\columnwidth/7\relax]{POEM-v2}%
\makebox[\dimexpr\columnwidth/7\relax]{UA-Fit}%
\makebox[\dimexpr\columnwidth/7\relax]{MAMMA}%
\makebox[\dimexpr\columnwidth/7\relax]{Ours$^*$}}\par\vspace{2pt}
\includegraphics[width=\columnwidth]{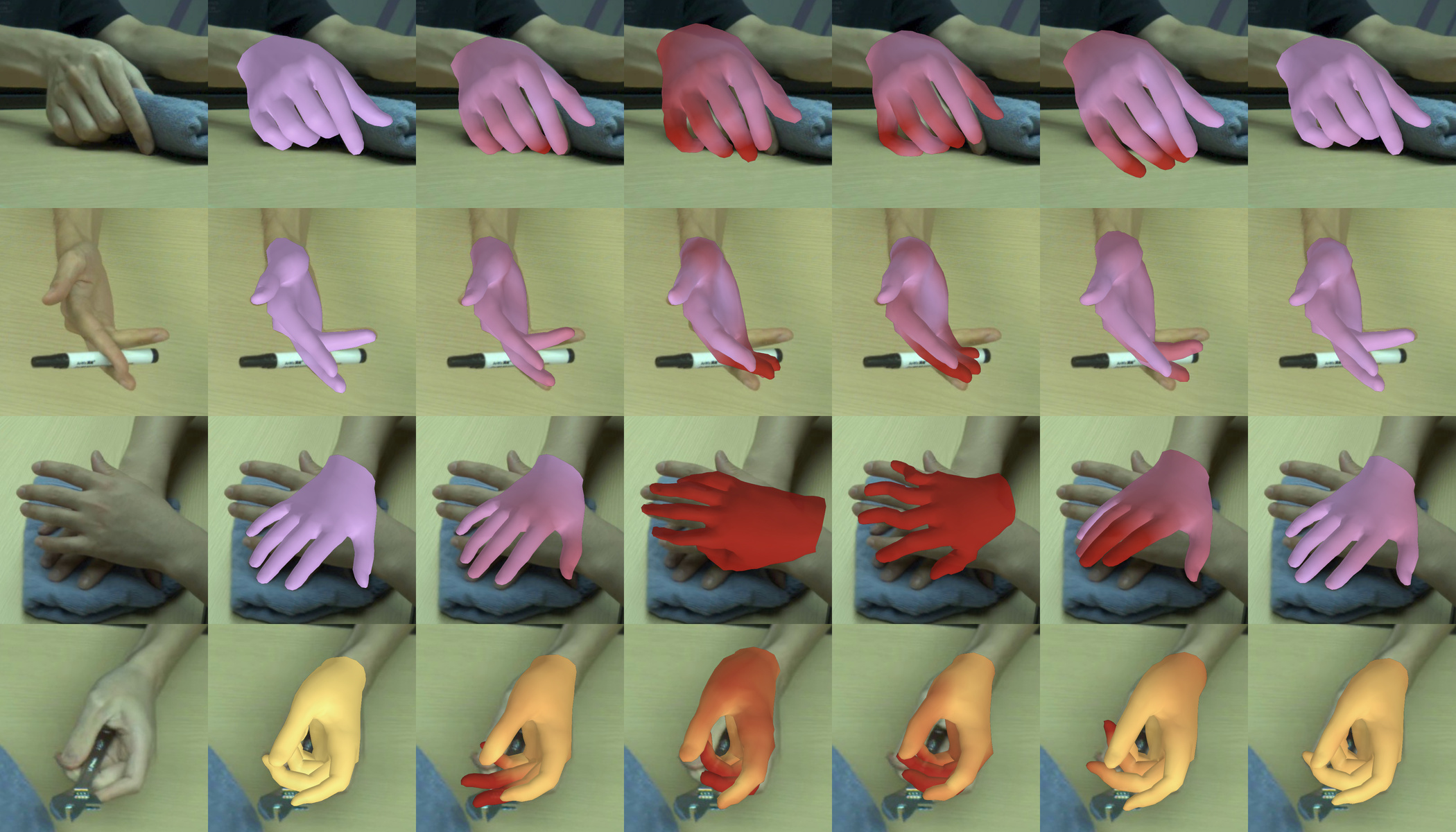}
\caption{Zero-shot results on HandQuery-Eval (eight-view input). Darker colors indicate larger per-vertex errors (0--20\,mm).}
\label{fig:qual_hqeval}
\end{figure}


\begin{figure}[t]
\centering
{\scriptsize
\makebox[9pt]{}%
\makebox[\dimexpr(\columnwidth-9pt)/7\relax]{Image}%
\makebox[\dimexpr(\columnwidth-9pt)/7\relax]{GT}%
\makebox[\dimexpr(\columnwidth-9pt)/7\relax]{WiLoR+Fit}%
\makebox[\dimexpr(\columnwidth-9pt)/7\relax]{POEM-v2}%
\makebox[\dimexpr(\columnwidth-9pt)/7\relax]{UA-Fit}%
\makebox[\dimexpr(\columnwidth-9pt)/7\relax]{MAMMA}%
\makebox[\dimexpr(\columnwidth-9pt)/7\relax]{Ours$^*$}}\par\vspace{1pt}
\makebox[9pt]{\rotatebox{90}{\makebox[\dimexpr(\columnwidth-9pt)*4/7\relax]{\scriptsize GigaHands}}}%
\includegraphics[width=\dimexpr\columnwidth-9pt\relax]{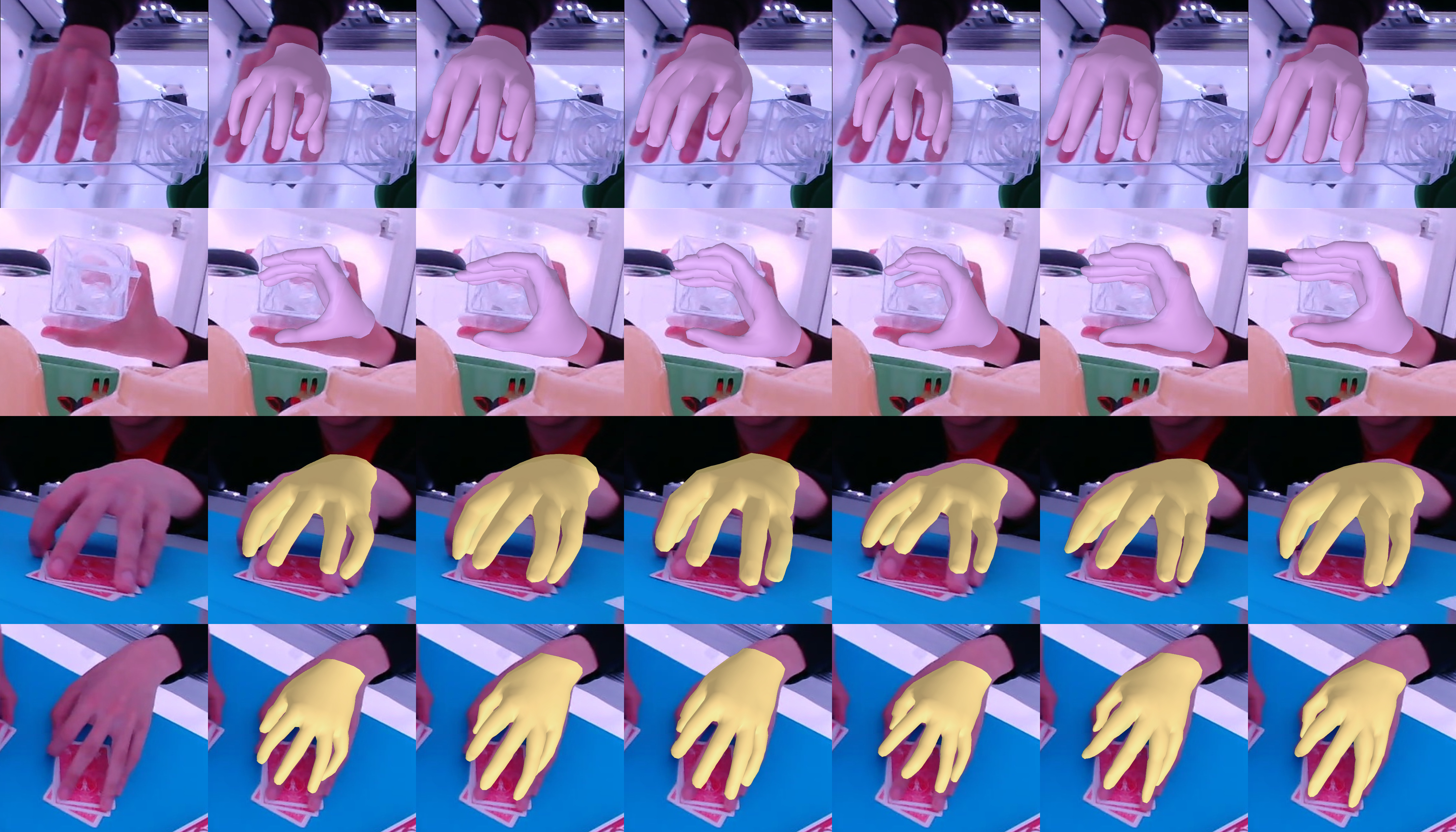}\par\vspace{2pt}
\makebox[9pt]{\rotatebox{90}{\makebox[\dimexpr(\columnwidth-9pt)*4/7\relax]{\scriptsize SHOW3D}}}%
\includegraphics[width=\dimexpr\columnwidth-9pt\relax]{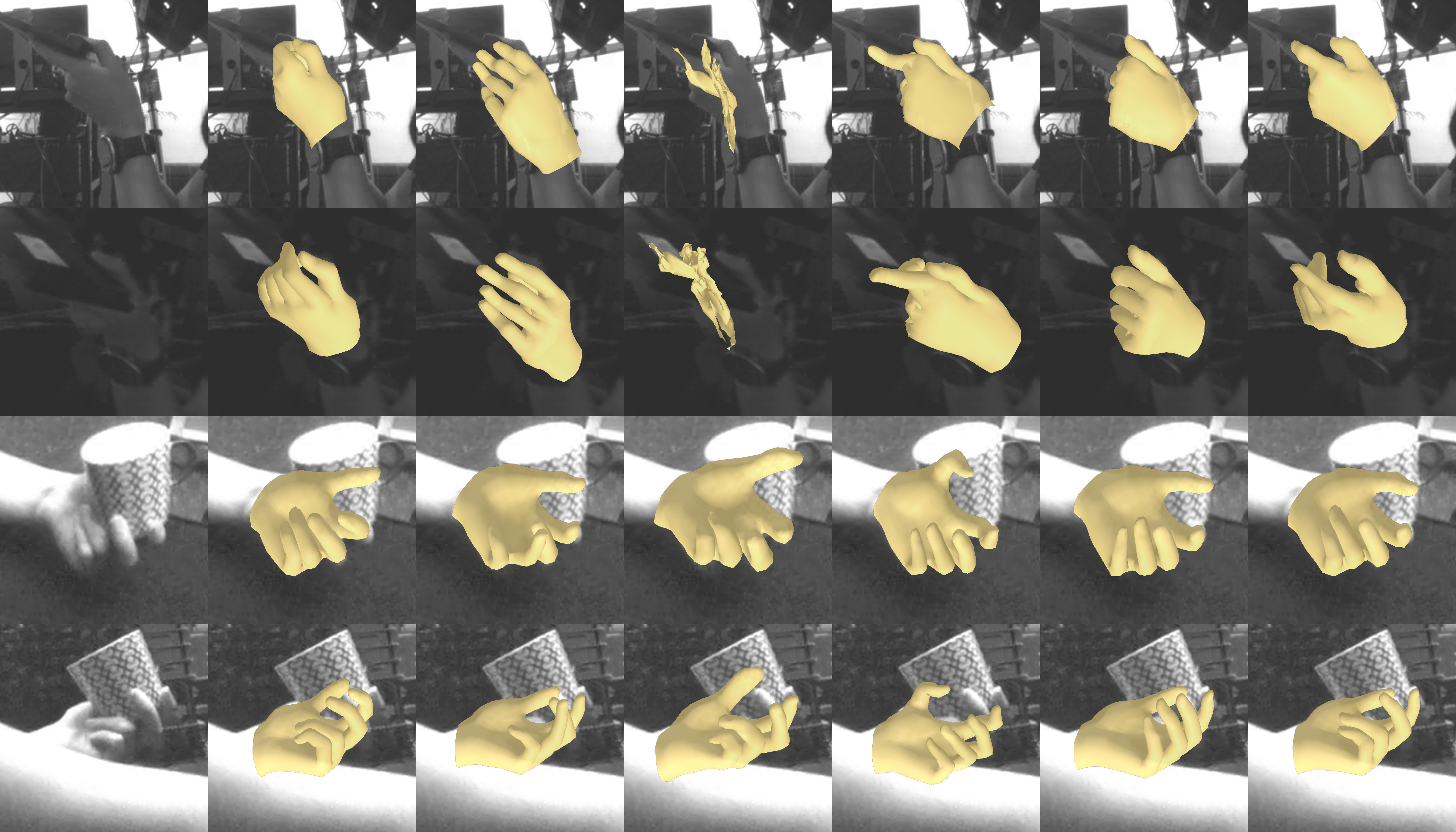}
\caption{Zero-shot results on GigaHands and SHOW3D. Four examples (two views each) with erroneous reference annotations. Left hands are shown in yellow and right hands in violet.}
\label{fig:qual_zs}
\end{figure}

\noindent\textbf{Qualitative results.}
Fig.~\ref{fig:qual_hqeval} compares per-vertex reconstruction errors on HandQuery-Eval, where our method exhibits fewer high-error regions than competing methods. We further evaluate on SHOW3D and GigaHands, whose capture conditions differ substantially from training. Since their reference hand annotations are generated automatically and can contain visible inaccuracies, we focus on qualitative comparisons. As shown in Fig.~\ref{fig:qual_zs}, our reconstructions better align with the observed images, particularly under challenging hand-object interactions and severe occlusions. These results provide further evidence of HandQuery's generalization to out-of-distribution capture settings. More qualitative comparisons are provided in Sec.~6 of the supplementary.

\subsection{Ablation Studies}
\label{sec:ablation}



\begin{figure}[t]
\centering
{\small%
\makebox[0.2\columnwidth]{Image}%
\makebox[0.2\columnwidth]{GT}%
\makebox[0.2\columnwidth]{DETR, 63}%
\makebox[0.2\columnwidth]{DETR, 778}%
\makebox[0.2\columnwidth]{Ours, 778}}\par\vspace{2pt}
\includegraphics[width=\columnwidth]{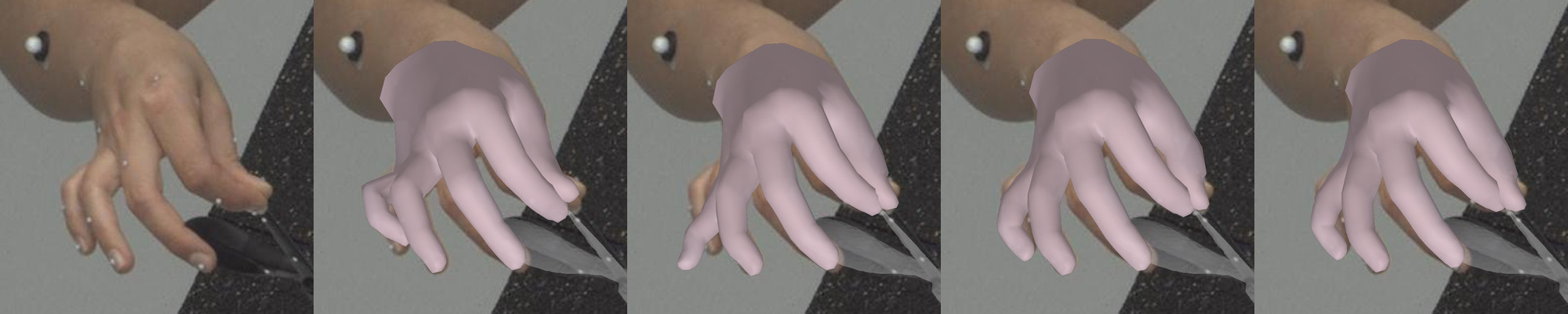}
\caption{Qualitative decoder ablation on ARCTIC-MV. DETR-63 misplaces the little finger, while DETR-778 and our independent-query decoder reconstruct it accurately.}
\label{fig:abl_decoder}
\end{figure}

\noindent\textbf{Decoder.}
We compare our independent query decoder with the self-attention-based decoder of MAMMA~\cite{cuevas2026mamma} (DETR variants). For a fair comparison, all variants share the same encoder, five real-world training datasets (without synthetic data), losses, and training schedule. We report MPVPE on the same hand-frames as Table~\ref{tab:exp_main}, along with relative training iteration time and peak per-GPU memory for an 8-view, 48-frame clip on NVIDIA A800 GPUs. Both DETR variants use activation checkpointing. As shown in Table~\ref{tab:exp_decoder}, DETR-778 achieves slightly better accuracy overall (by at most 0.2\,mm), but incurs substantially higher training time and memory costs. Reducing DETR to 63 queries lowers its overhead, yet it remains slower and more memory-intensive than ours and struggles with challenging cases, as shown in Fig.~\ref{fig:abl_decoder}. In contrast, our lightweight decoder enables training on GPUs with more modest memory capacity. Moreover, incorporating synthetic data (Ours*) doubles the total training time while achieving the best overall performance. Training DETR-778 with the same additional data would instead require an estimated 6.4$\times$ the training time of Ours (approximately 96 hours). These trade-offs motivate our choice of independent vertex queries.

\begin{table}[t]
\centering
\footnotesize
\setlength{\tabcolsep}{0pt}
\setlength{\abovecaptionskip}{4pt}
\setlength{\aboverulesep}{0pt}
\setlength{\belowrulesep}{0pt}
\renewcommand{\arraystretch}{1.22}
\caption{Decoder ablation. Time denotes relative training iteration time (Ours: $1.0\times$), and Mem. denotes peak GPU memory (GB).}

\begin{tabular*}{\columnwidth}{@{\hspace{1pt}}l@{\hspace{4pt}}c@{\hspace{5pt}\vrule width \arrayrulewidth\extracolsep{\fill}}c@{\hspace{3pt}}c@{\hspace{3pt}}c@{\hspace{3pt}}c@{\hspace{3pt}}c@{}c@{\hspace{3pt}\vrule width \arrayrulewidth\hspace{3pt}}c@{\hspace{4pt}}c@{\hspace{1pt}}}
\toprule
 & & \multicolumn{5}{c}{MPVPE (mm)} & & \multicolumn{2}{c@{\hspace{1pt}}}{Training} \\
\cmidrule{3-7}\cmidrule{9-10}
Decoder & Pts & DexYCB & HO3D & OakInk & ARCTIC & IH & & Time & \multicolumn{1}{c@{\hspace{1pt}}}{Mem.} \\
\midrule
DETR & 63 & 4.63 & 6.16 & 5.49 & 4.86 & 3.91 &  & 1.1$\times$ & 47.4 \\
DETR & 778 & 4.60 & \textbf{6.08} & 5.48 & \textbf{4.71} & \textbf{3.76} &  & 3.2$\times$ & 78.4 \\
\midrule
Ours & 778 & \textbf{4.56} & 6.11 & \textbf{5.47} & 4.85 & 3.96 &  & \textbf{1.0$\times$} & \textbf{39.3} \\
\bottomrule
\end{tabular*}

\label{tab:exp_decoder}
\end{table}

\begin{figure}[t]
\centering
\includegraphics[width=\columnwidth]{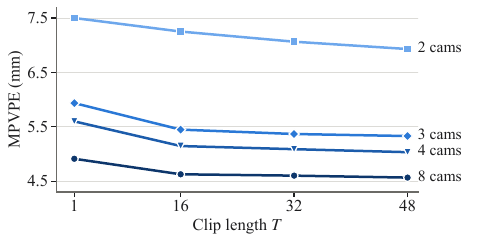}
\caption{MPVPE versus clip length on DexYCB-MV with varying camera counts. All variants are trained on the five real datasets without synthetic data; $T=48$ corresponds to Ours in Table~\ref{tab:exp_main}.}
\label{fig:exp_tlen}
\end{figure}

\noindent\textbf{Clip length.}
We train variants with clip lengths $T\in\{1,16,32,48\}$, adjusting the batch size to maintain 384 frames per iteration. $T=1$ omits temporal attention. Evaluation uses length-matched sliding windows with stride $\max(1,T/2)$. As shown in Fig.~\ref{fig:exp_tlen}, temporal context consistently improves reconstruction accuracy across all camera counts, reducing MPVPE by 7.1--10.2\% from $T=1$ to $T=48$. With three or more cameras, most improvements are achieved by $T=16$. In contrast, with only two cameras, extending the clip from 16 to 48 frames further reduces MPVPE from 7.25 to 6.93\,mm (4.4\%). This suggests that longer temporal context is particularly beneficial under sparse camera coverage, where observations from additional frames can help compensate for missing viewpoints.


\section{Conclusion}
\label{sec:conclusion}

We introduced HandQuery for 4D hand reconstruction from synchronized and calibrated multi-view videos. Its central design combines multi-view spatiotemporal feature fusion with independent vertex queries, allowing landmark prediction to draw on complementary observations while supporting flexible vertex selection. Our synthetic multi-view corpus provides consistent dense supervision under diverse observation conditions. Evaluations on five real-world benchmarks and a newly captured studio dataset demonstrate strong reconstruction accuracy, robustness to fewer cameras, and zero-shot generalization. These results highlight the value of fusing complementary observations before landmark prediction while retaining explicit surface correspondences for geometric reconstruction. 

A remaining limitation is our reliance on hand detection for input cropping. Future directions include detection-free reconstruction and joint hand-object modeling to better leverage interaction context.
\section*{LLM Usage Statement}
The core research idea originated from the authors. OpenAI's ChatGPT was used to assist with subsequent experimental development, iterative method refinement, and manuscript polishing. In constructing the HandQuery-Real dataset, LLM-based agents assisted in identifying and repairing annotation errors, with all frames subsequently inspected and verified by human annotators. All methodological decisions, experimental results, and final annotations were reviewed and validated by the authors, who take full responsibility for the scientific content of this work.
\clearpage
\setcounter{page}{1}
\setcounter{section}{0}
\setcounter{figure}{0}
\setcounter{table}{0}
\setcounter{equation}{0}
\renewcommand{\thefigure}{S\arabic{figure}}
\renewcommand{\thetable}{S\arabic{table}}
\renewcommand{\theequation}{S\arabic{equation}}
\renewcommand{\topfraction}{0.90}
\renewcommand{\bottomfraction}{0.85}
\renewcommand{\textfraction}{0.08}
\renewcommand{\dbltopfraction}{0.90}
\renewcommand{\dblfloatpagefraction}{0.50}
\setcounter{dbltopnumber}{2}
\maketitlesupplementary

\noindent\textbf{Supplementary Overview}. This supplementary material is organized as follows. Sec.~\ref{appendix:implementation} provides implementation details and evaluation protocols. Sec.~\ref{appendix:synthetic_dataset} describes the motivation and generation pipeline of HandQuery-Syn. Sec.~\ref{appendix:real_dataset} introduces HandQuery-Real and HandQuery-Eval, including annotation procedures and optical motion-capture validation. Sec.~\ref{appendix:more_experiments} analyzes inference efficiency and cross-view fusion, while Sec.~\ref{appendix:temporal} examines temporal consistency. Sec.~\ref{appendix:qualitative} presents additional qualitative comparisons on unseen datasets. Sec.~\ref{appendix:retargeting} demonstrates dexterous robot hand retargeting as a downstream application, and Sec.~\ref{appendix:ethics} concludes with an ethics statement.

\section{Implementation Details}
\label{appendix:implementation}

\subsection{Architecture and Perspective Cropping}
\label{appendix:architecture}

\noindent\textbf{Perspective crop cameras.}
Given a hand bounding box in view $c$ at frame $t$ with side length $b_{c,t} = \max(x_1 - x_0, y_1 - y_0)$ and center $\mathbf{p}_{c,t}$, we construct a virtual pinhole camera rotated so that its optical axis points along $\mathbf{K}_c^{-1}[\mathbf{p}_{c,t}, 1]^\top$ while preserving the vertical orientation of the physical camera. Each crop spans $1.4\times$ the maximum box side within that view's clip at resolution $S \times S$ with $S = 256$, keeping the virtual focal length constant across the $T$ frames while centering the hand. The resulting virtual intrinsics $\mathbf{K}^{\mathrm{crop}}_{c,t}$ and extrinsics $\mathbf{E}^{\mathrm{crop}}_{c,t}$ are used both for PRoPE~\cite{li2026cameras} and for mapping predicted landmarks back to the original image plane.

\noindent\textbf{Multi-view spatiotemporal encoder.}
We use a 12-layer ViT-B~\cite{dosovitskiy2021an} backbone with patch size $16$, feature dimension $D = 768$, $12$ attention heads, MLP ratio $4.0$, and drop-path rate $0.1$, producing $P = 256$ patch tokens per crop. A fixed sinusoidal temporal positional encoding of the within-clip frame index is added to all patch tokens before the first block. Following TimeSformer~\cite{gberta_2021_ICML}, each block first applies divided temporal self-attention across the $T$ frames at a fixed view and patch index, with zero-initialized output projections so training starts from the pretrained single-image representation. Following Depth Anything 3~\cite{lin2026depth}, within-image and cross-view interactions share the ViT self-attention weights through token regrouping: the first four blocks apply spatial attention within each crop, whereas the remaining eight blocks alternate between cross-view attention over all $V \times P$ tokens at frame $t$ and within-image spatial attention, matching the $1:2$ stage ratio in Sec.~\ref{subsection:encoder} of the main paper. Within cross-view attention, PRoPE~\cite{li2026cameras} encodes focal-normalized crop intrinsics together with camera translations normalized by the mean camera-to-hand distance of the clip.

\noindent\textbf{Independent dense vertex query decoder.}
We define $778$ learned query embeddings per hand in $\mathbb{R}^{768}$, initialized from $\mathcal{N}(0, 1)$ and combined with the same sinusoidal temporal encoding before entering six pre-norm cross-attention and feed-forward blocks. Each query attends solely to the target image memory $\mathbf{Z}_{c,t}$ without query-to-query self-attention, as described in Sec.~\ref{subsection:decoder} of the main paper. The complete network has $157.9\mathrm{M}$ parameters, comprising $114.2\mathrm{M}$ in the encoder and $43.7\mathrm{M}$ in the decoder and prediction heads. Training samples $64$ labeled vertex queries per view and frame, whereas inference decodes all $778$ vertices of each target hand.

\subsection{Training Objectives}
\label{appendix:losses}

For each queried vertex $(i, c, t)$ with normalized ground-truth crop coordinate $\mathbf{u}_{i,c,t} \in [-1, 1]^2$, prediction $\hat{\mathbf{u}}_{i,c,t}$, and clamped log-uncertainty $\ell_{i,c,t} = \mathrm{clamp}(\log \hat{\sigma}_{i,c,t}, -3, 6)$ in crop pixels, let $d^2_{i,c,t} = \|\hat{\mathbf{u}}_{i,c,t} - \mathbf{u}_{i,c,t}\|_2^2 \cdot (S/2)^2$ denote the squared pixel error. Following MAMMA~\cite{cuevas2026mamma}, we adopt a capped Gaussian negative log-likelihood and introduce an additional weak, uncertainty-independent Euclidean-distance penalty:
{\setlength{\abovedisplayskip}{6pt}\setlength{\belowdisplayskip}{6pt}%
\begin{equation}
\begin{aligned}
  \mathcal{L}_{\mathrm{pos}}
  &= \frac{1}{|\Omega_{\mathrm{pos}}|}
     \sum_{(i,c,t)\in\Omega_{\mathrm{pos}}}
     \bigg[
       \min\!\left(\frac{d^2_{i,c,t}}{2 e^{2\ell_{i,c,t}}},\, \tau\right) \\
  &\qquad + 2\ell_{i,c,t} + \lambda_{d}\sqrt{d^2_{i,c,t} + \epsilon}\,
     \bigg],
\end{aligned}
  \label{eq:suppl_lpos}
\end{equation}}%
where $\tau = 25$, $\lambda_{d} = 0.05$, $\epsilon = 10^{-6}$, and $\Omega_{\mathrm{pos}}$ collects labeled vertices in front of the crop camera with $\|\mathbf{u}_{i,c,t}\|_\infty < 2.0$. The additional penalty maintains direct localization supervision when the likelihood term is capped or down-weighted by high predicted uncertainty. Both visible and occluded vertices receive 2D supervision whenever reference meshes are available, encouraging the encoder to disambiguate occluded regions from complementary views and frames. Visibility logits are supervised with binary cross-entropy on vertices with valid visibility labels. For Ours$^*$, the hand--object and hand--hand contact heads are additionally supervised on synthetic clips using sigmoid focal loss with $\alpha = 0.9$ and $\gamma = 2.0$, with unit loss weight on all auxiliary heads.

\subsection{Training Protocol and Optimization}
\label{appendix:training}

\noindent\textbf{Initialization.}
The ViT-B encoder is initialized from ViTPose-B~\cite{xu2022vitpose}, with its spatial positional embeddings bicubic-interpolated from $16\times 12$ to $16\times 16$ to match our $256\times 256$ crops. The temporal attention layers and vertex query decoder are initialized from scratch.

\noindent\textbf{Multi-view video training.}
We train the network end-to-end in bf16 mixed precision using AdamW with weight decay $0.05$ and gradient norm clipping at $1.0$. Each iteration processes $8$ multi-view clips of up to $T = 48$ frames and $V \in [2, 8]$ randomly sampled cameras; when both hands appear in a crop, visible vertices of the second hand are supervised jointly. The peak learning rate is set to $2\times 10^{-5}$ for the encoder and $1\times 10^{-4}$ for the temporal layers and decoder, with $1{,}000$ linear warmup steps followed by cosine decay to $1\%$ of the peak rate. Real-only training runs for $40\mathrm{k}$ steps on DexYCB, HO3D, OakInk, ARCTIC, and $30$\,fps InterHand2.6M with sampling weights $0.25$, $0.10$, $0.20$, $0.25$, and $0.20$. Training with HandQuery-Syn runs for $80\mathrm{k}$ steps, sampling synthetic and real clips in equal proportion while preserving the relative weights among the five real datasets.

\subsection{Model Fitting}
\label{appendix:model_fitting}

\noindent\textbf{Sliding-window inference and initialization.}
At test time, sequences are processed with overlapping windows of length $T = 48$ and stride $24$. Crop-space landmark predictions are mapped back to full-image coordinates via the inverse crop homography, scaling $\hat{\sigma}$ by the local Jacobian, and overlapping window estimates are fused by inverse-variance weighting $1/\hat{\sigma}^2$ with $\sigma_{c,t,i}$ clamped to $[1, 50]$\,px. Vertices predicted as visible with $p_{c,t,i} > 0.5$ in at least two views are triangulated via inverse-variance weighted DLT to obtain 3D points $\tilde{\mathbf{X}}_{t,i}$, discarding outliers farther than $15$\,cm from the per-frame median. Frames with at least $10$ valid triangulated vertices initialize global orientation $\boldsymbol{\theta}^{\mathrm{glob}}_t \in \mathbb{R}^3$ and translation $\mathbf{t}_t \in \mathbb{R}^3$ via similarity Procrustes alignment from the neutral MANO template, while remaining frames propagate initialization from neighboring frames.

\noindent\textbf{Three-stage MANO optimization.}
We optimize per-frame $\boldsymbol{\theta}^{\mathrm{glob}}_t$, $45$-DoF PCA pose $\boldsymbol{\alpha}_t$, translation $\mathbf{t}_t$, and sequence-shared shape $\boldsymbol{\beta} \in \mathbb{R}^{10}$ using Adam across three stages of $150$, $300$, and $200$ steps, with learning rate $0.01$ in the first two stages and $0.005$ in the third. Given MANO vertex $\mathbf{X}_{t,i}$ and its projection $\pi_c(\mathbf{X}_{t,i})$ in camera $c$ at depth $z_{c,t,i}$, we combine a Geman--McClure reprojection term $\rho_\delta(r) = r^2/(\delta^2 + r^2)$ with $\delta = 8$, a 3D triangulation anchor, and a pose prior:
{\setlength{\abovedisplayskip}{6pt}\setlength{\belowdisplayskip}{6pt}%
\begin{alignat}{2}
  &\mathcal{L}_{\mathrm{rep}}  &&{}= \frac{\sum_{c,t,i} \omega_{c,t,i}\,\rho_\delta\!\left(\frac{\|\pi_c(\mathbf{X}_{t,i}) - \mathbf{u}_{c,t,i}\|_2}{\sigma_{c,t,i}}\right)}{\sum_{c,t,i} \omega_{c,t,i}},
  \label{eq:suppl_fit_rep} \\
  &\mathcal{L}_{\mathrm{3D}}   &&{}= \frac{\sum_{t,i} o_{t,i}\,\|\mathbf{X}_{t,i} - \tilde{\mathbf{X}}_{t,i}\|_2}{\sum_{t,i} o_{t,i}},
  \label{eq:suppl_fit_3d} \\
  &\mathcal{L}_{\mathrm{pose}} &&{}= \frac{1}{45 T}\sum_{t=1}^{T}\|\boldsymbol{\alpha}_t\|_2^2,
  \label{eq:suppl_fit_pose}
\end{alignat}}%
where $\omega_{c,t,i} = \max(p_{c,t,i}, p_{\min})\,m_{c,t}\,\mathbb{I}[z_{c,t,i} > 0.05\text{\,m}]$ uses $p_{\min} = 0.8$ in Stage~1 and $0.2$ thereafter, $m_{c,t}$ indicates valid views, and $o_{t,i}$ marks valid triangulated points. Stage~1 updates only $\{\boldsymbol{\theta}^{\mathrm{glob}}_t, \mathbf{t}_t\}$ under $5\mathcal{L}_{\mathrm{rep}} + 100\mathcal{L}_{\mathrm{3D}}$; Stage~2 optimizes all parameters under $20\mathcal{L}_{\mathrm{rep}} + 20\mathcal{L}_{\mathrm{3D}} + 0.02\mathcal{L}_{\mathrm{pose}}$; and Stage~3 reduces the 3D anchor weight to $5$ for fine image alignment. No temporal smoothness or shape regularization penalty is used.

\subsection{Benchmark Splits and Baseline Protocols}
\label{appendix:benchmarks}

\noindent\textbf{Five public benchmarks.}
We follow the multi-view evaluation protocols of POEM and POEM-v2~\cite{yang2023poem,yang2025multi}: the official right-hand S0 test split on eight-camera DexYCB-MV~\cite{chao2021dexycb}; held-out sequences GPMF1 and SB1 on five-camera HO3D-MV~\cite{hampali2020honnotate}; every fourth test frame of object split SP2 on four-camera OakInk-MV~\cite{yang2022oakink}; right-hand P1 validation sequences visible across all eight allocentric views on ARCTIC-MV~\cite{fan2023arctic}; and $30$\,fps validation sequences evaluated across $17$ fixed eight-camera groups under POEM-v2's ground truth on InterHand-MV~\cite{moon2020interhand2}. POEM-v2 and UA-Fit are omitted on OakInk-MV because their released models adopt the subject split, which overlaps the SP2 test sequences. On InterHand-MV, recomputing crop centers and scales from in-image joints corrects enlarged boxes caused by out-of-frame joints in POEM-v2's evaluation package, reducing POEM-v2's MPVPE from $7.62$ to $5.92$\,mm.

\begin{figure*}[tp]
  \centering
  {\small%
  \makebox[15pt]{}%
  \makebox[0.1826\textwidth]{ARCTIC}\hspace{0.0067\textwidth}%
  \makebox[0.1826\textwidth]{DexYCB}\hspace{0.0067\textwidth}%
  \makebox[0.1826\textwidth]{HO3D}\hspace{0.0067\textwidth}%
  \makebox[0.1826\textwidth]{OakInk}\hspace{0.0067\textwidth}%
  \makebox[0.1826\textwidth]{InterHand2.6M}}\par\vspace{2pt}
  \leavevmode\vbox to 0.3686\textwidth{\offinterlineskip
    \vbox to 0.1826\textwidth{\vfil\hbox to 13pt{\hfil\rotatebox{90}{\footnotesize Input}\hfil}\vfil}%
    \vskip 0.0034\textwidth
    \vbox to 0.1826\textwidth{\vfil\hbox to 13pt{\hfil\rotatebox{90}{\footnotesize Annotation}\hfil}\vfil}%
  }\hspace{2pt}%
  \includegraphics[width=0.94\textwidth]{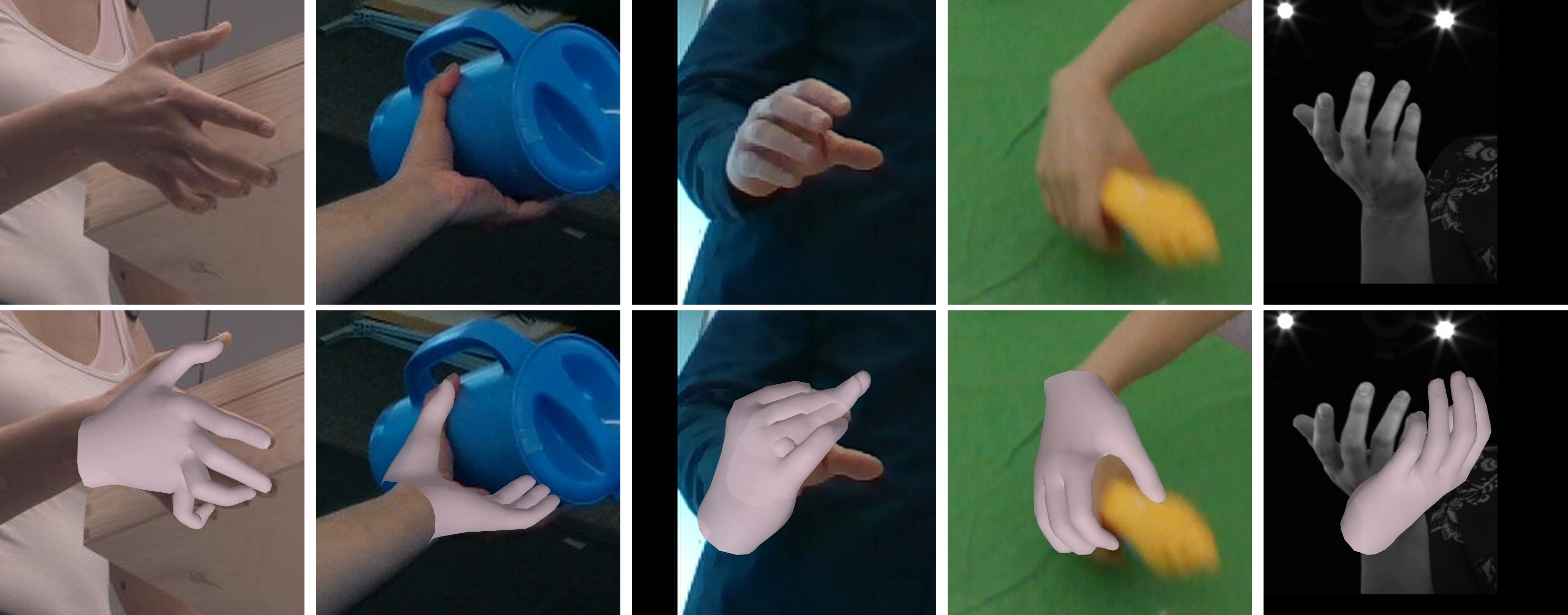}\par\vspace{3pt}
  \leavevmode\vbox to 0.3686\textwidth{\offinterlineskip
    \vbox to 0.1826\textwidth{\vfil\hbox to 13pt{\hfil\rotatebox{90}{\footnotesize Input}\hfil}\vfil}%
    \vskip 0.0034\textwidth
    \vbox to 0.1826\textwidth{\vfil\hbox to 13pt{\hfil\rotatebox{90}{\footnotesize Annotation}\hfil}\vfil}%
  }\hspace{2pt}%
  \includegraphics[width=0.94\textwidth]{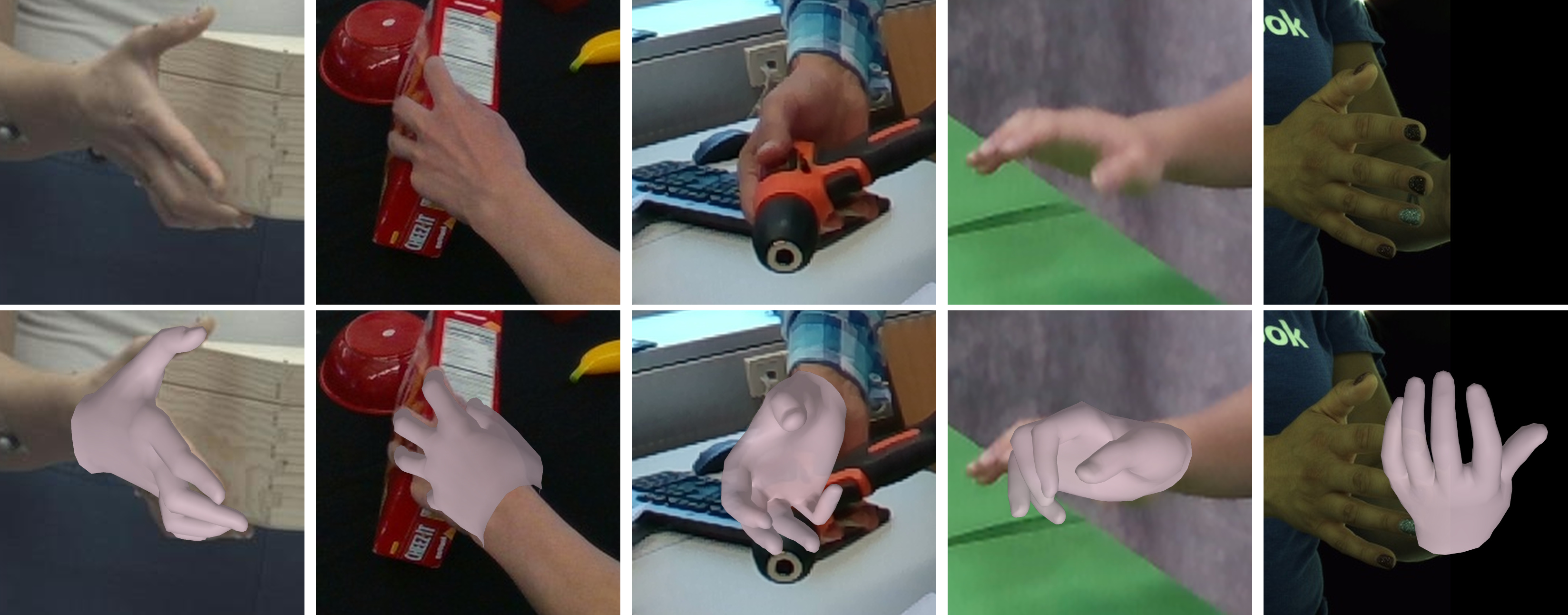}\par\vspace{3pt}
  \leavevmode\vbox to 0.3686\textwidth{\offinterlineskip
    \vbox to 0.1826\textwidth{\vfil\hbox to 13pt{\hfil\rotatebox{90}{\footnotesize Input}\hfil}\vfil}%
    \vskip 0.0034\textwidth
    \vbox to 0.1826\textwidth{\vfil\hbox to 13pt{\hfil\rotatebox{90}{\footnotesize Annotation}\hfil}\vfil}%
  }\hspace{2pt}%
  \includegraphics[width=0.94\textwidth]{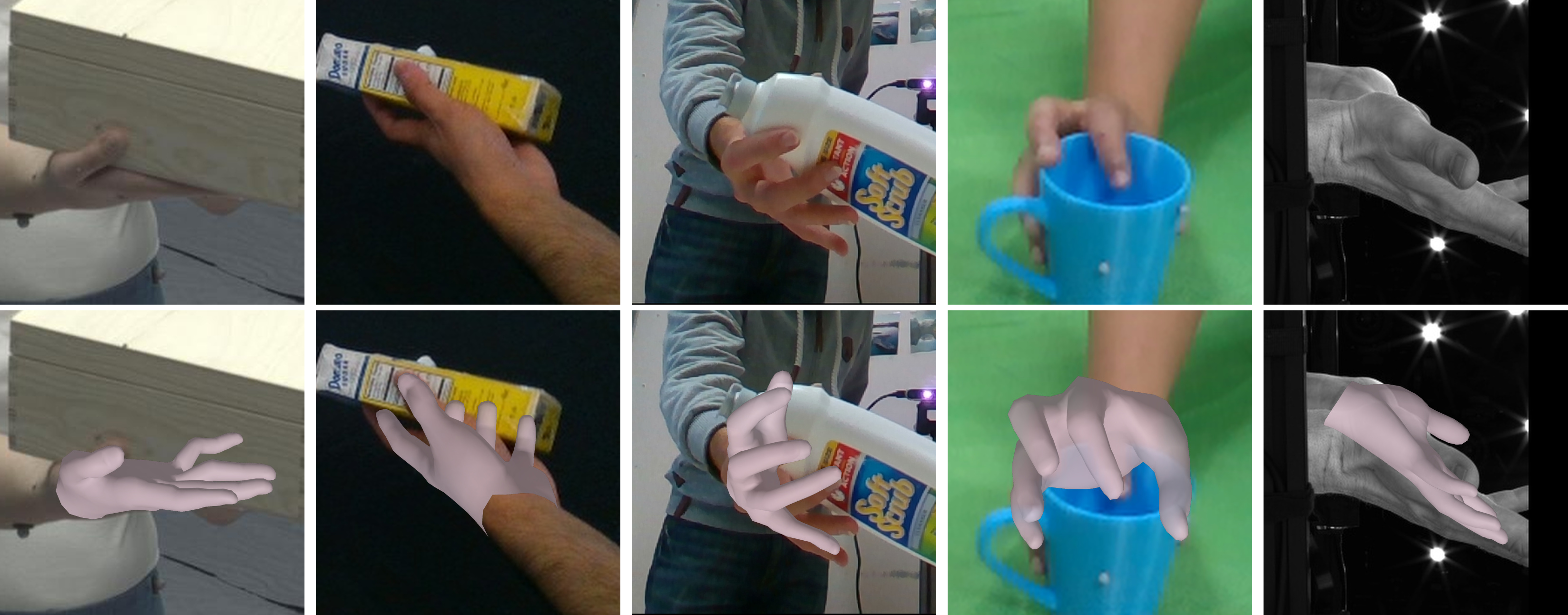}
  \caption{Examples of annotation misalignment across the five real training benchmarks. Each vertical pair shows an input crop above the corresponding MANO annotation projected into the same view. Published annotations can exhibit rotated palms, collapsed digits, or spatial offsets relative to the observed hand.}
  \label{fig:suppl_misalign}
\end{figure*}

\noindent\textbf{Single-view + fit baselines.}
For WiLoR~\cite{potamias2025wilor} + Fit, we extract $2.0\times$ crops around the ground-truth boxes, project the predicted $778$ vertices back into each view, and apply our multi-view MANO fitter with uniform visibility and an uncertainty scale calibrated on DexYCB validation. For Sapiens2~\cite{khirodkar2026sapiens2} + Fit, we use $2.5\times$ square crops and fit to the $21$ predicted keypoints with confidence at least $0.3$. For MAMMA~\cite{cuevas2026mamma}, we retrain its ViT-B and six-layer DETR architecture from the same ViTPose-B initialization for $40\mathrm{k}$ steps on the exact real-and-synthetic mixture used by Ours$^*$, expanding its landmark vocabulary via farthest-point sampling to $256$ queries per hand and reconstructing meshes with our fitter, which outperforms MAMMA's original L-BFGS fitter across all five benchmarks.

\section{HandQuery-Syn}
\label{appendix:synthetic_dataset}

\subsection{Annotation Misalignment in Real Datasets}
\label{appendix:misalignment}

\begin{figure*}[t]
  \centering
  {\small%
  \makebox[16pt]{}%
  \makebox[0.1888\textwidth]{$t_1$}\hspace{0.0052\textwidth}%
  \makebox[0.1888\textwidth]{$t_2$}\hspace{0.0052\textwidth}%
  \makebox[0.1888\textwidth]{$t_3$}\hspace{0.0052\textwidth}%
  \makebox[0.1888\textwidth]{$t_4$}\hspace{0.0052\textwidth}%
  \makebox[0.1888\textwidth]{$t_5$}}\par\vspace{2pt}
  \leavevmode\vbox to 0.1416\textwidth{\vfil\hbox to 14pt{\hfil\rotatebox{90}{\footnotesize ARCTIC}\hfil}\vfil}\hspace{2pt}%
  \includegraphics[width=0.965\textwidth]{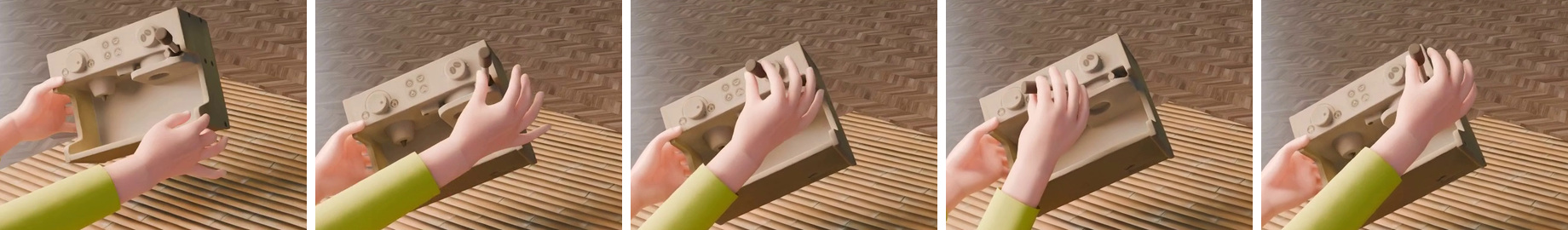}\par\vspace{2pt}
  \leavevmode\vbox to 0.1416\textwidth{\vfil\hbox to 14pt{\hfil\rotatebox{90}{\footnotesize DexYCB}\hfil}\vfil}\hspace{2pt}%
  \includegraphics[width=0.965\textwidth]{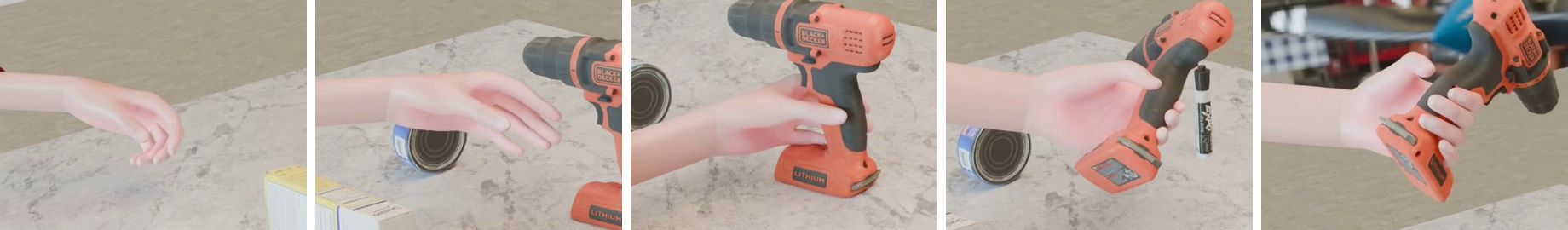}\par\vspace{2pt}
  \leavevmode\vbox to 0.1416\textwidth{\vfil\hbox to 14pt{\hfil\rotatebox{90}{\footnotesize OakInk}\hfil}\vfil}\hspace{2pt}%
  \includegraphics[width=0.965\textwidth]{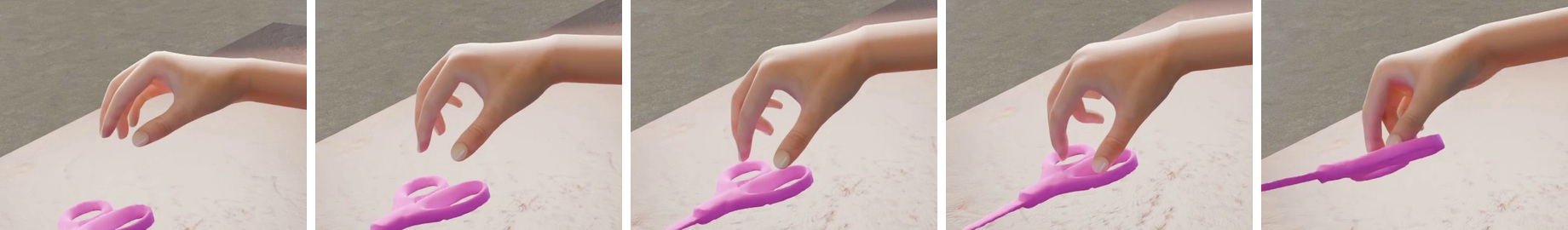}\par\vspace{2pt}
  \leavevmode\vbox to 0.1416\textwidth{\vfil\hbox to 14pt{\hfil\rotatebox{90}{\footnotesize HO3D}\hfil}\vfil}\hspace{2pt}%
  \includegraphics[width=0.965\textwidth]{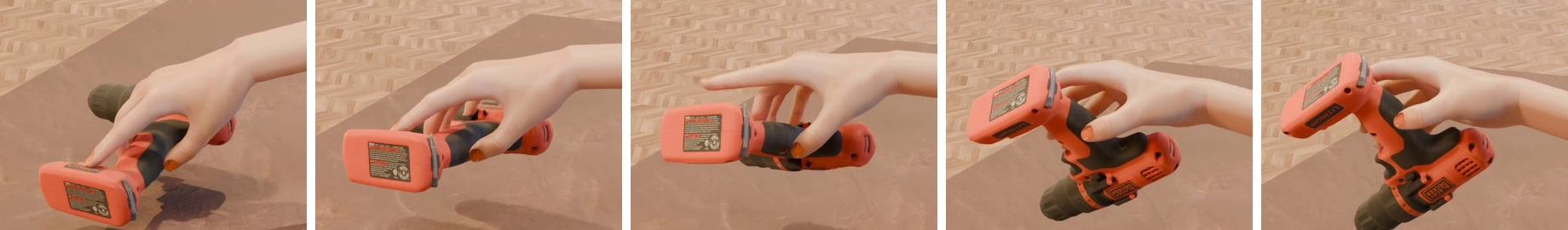}\par\vspace{2pt}
  \leavevmode\vbox to 0.1416\textwidth{\vfil\hbox to 14pt{\hfil\rotatebox{90}{\footnotesize InterHand2.6M}\hfil}\vfil}\hspace{2pt}%
  \includegraphics[width=0.965\textwidth]{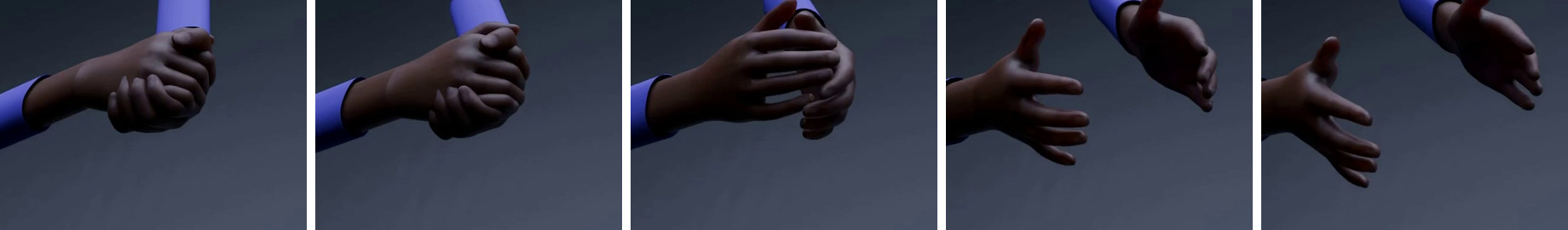}
  \caption{Examples from HandQuery-Syn across the five motion sources. Each row shows crops from five timestamps of a rendered sequence, illustrating varied articulations, skin and sleeve appearances, tabletop materials, and lighting.}
  \label{fig:suppl_syn}
\end{figure*}

As discussed in Sec.~\ref{subsection:handquery_sync} of the main paper, real multi-view hand benchmarks rely on multi-stage calibration and fitting pipelines in which synchronization drift, occlusion ambiguity, and automated fitting errors can leave reference meshes locally misaligned with the images. Fig.~\ref{fig:suppl_misalign} shows representative examples across all five training benchmarks where projected MANO annotations deviate from the observed palm orientation or finger articulation. When projected into dense 2D vertex targets, these discrepancies introduce conflicting supervision across views and frames. Re-rendering 3D motions in simulation guarantees exact geometric consistency among rendered pixels, camera parameters, vertex visibility, and surface contact labels.

\subsection{Synthetic Generation Pipeline}
\label{appendix:syn_pipeline}

\begin{figure*}[t]
  \centering
  \includegraphics[width=\textwidth]{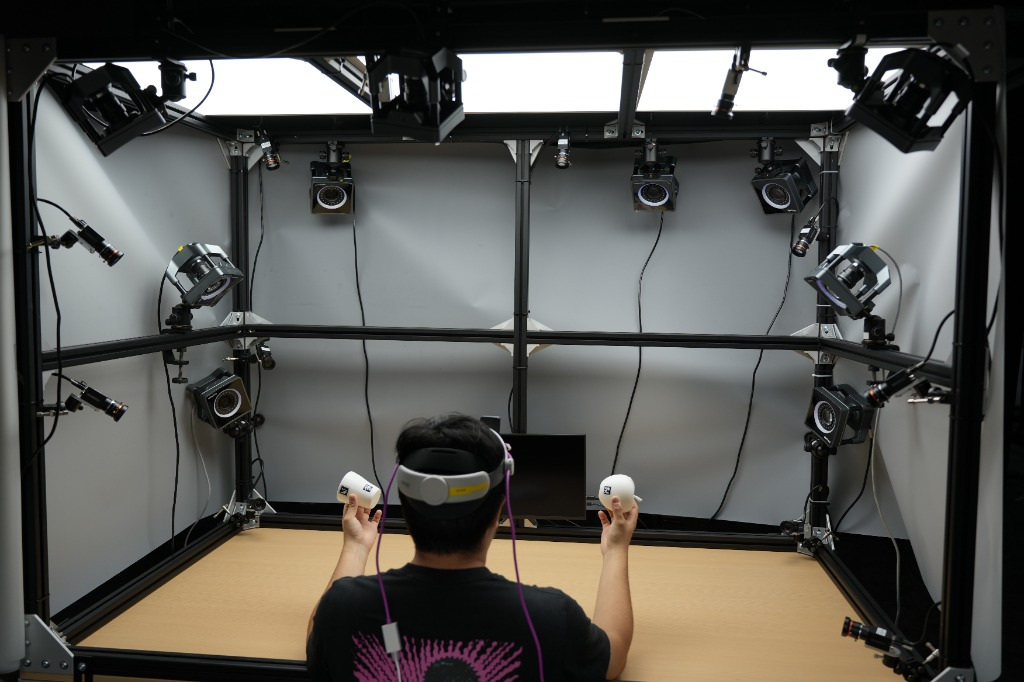}
  \caption{Multi-view tabletop capture rig integrating $12$ calibrated RGB cameras and $10$ optical motion-capture cameras, hardware-synchronized via PTP with sub-$50\,\mu\mathrm{s}$ exposure skew.}
  \label{fig:suppl_capture_rig}
  \vspace{-4pt}
\end{figure*}

We construct HandQuery-Syn in Blender Cycles from real 3D hand and object trajectories, generating $10\mathrm{k}$ multi-view scenes across the five motion sources (Fig.~\ref{fig:suppl_syn}).

\noindent\textbf{Motion curation.}
MANO and object trajectories are taken exclusively from the training splits of DexYCB, HO3D, OakInk, ARCTIC, and InterHand2.6M, strictly excluding all validation and test splits, and transformed into a canonical metric tabletop coordinate frame. Candidate clips of up to $60$ frames at $30$\,fps are retained only when at least $95\%$ of frames carry valid annotations and, for hand--object sequences, at least half of the frames exhibit hand--object contact within $1$\,cm so idle intervals are removed.

\noindent\textbf{Appearance and scene composition.}
Following AnyHand~\cite{si2026anyhand}, procedural forearms with randomized sleeves and clothing are attached to the MANO meshes to provide realistic wrist context and occlusions. Skin appearance combines subsurface scattering with $270$ UV-space DART textures~\cite{gao2022dart} spanning diverse skin tones, nails, blemishes, and tattoos. Tabletops and floors sample PBR materials across wood, stone, plastic, and fabric categories, illuminated by randomized indoor HDRI environments and area lights.

\noindent\textbf{Camera sampling and dense labels.}
A perturbed $32$-camera dome is placed around the workspace center. From viewpoints above the tabletop with at least $80\%$ in-frame hand coverage, we draw one view at random and select seven additional views via farthest-point sampling, rendering eight synchronized views per scene at $1280\times 960$ resolution with OptiX denoising. Exact camera projection, ray-cast occlusion testing, and $1$\,cm surface proximity directly yield all $778$ 2D vertex coordinates, binary visibility labels, and hand--object and hand--hand contact annotations.

\section{HandQuery-Real and HandQuery-Eval}
\label{appendix:real_dataset}

\subsection{Capture Rig and Evaluation Subset}
\label{appendix:real_setup}

\begin{figure*}[t]
  \centering
  {\small%
  \makebox[16pt]{}%
  \makebox[0.1787\textwidth]{Left Input}\hspace{0.0048\textwidth}%
  \makebox[0.1787\textwidth]{Left Ours$^*$}\hspace{0.0048\textwidth}%
  \makebox[0.1787\textwidth]{Right Input}\hspace{0.0048\textwidth}%
  \makebox[0.1787\textwidth]{Right Ours$^*$}\hspace{0.0060\textwidth}%
  \makebox[0.2299\textwidth]{3D Reconstruction}}\par\vspace{2pt}
  \leavevmode\vbox to 0.1656\textwidth{\vfil\hbox to 14pt{\hfil\rotatebox{90}{\footnotesize Interlocking}\hfil}\vfil}\hspace{2pt}%
  \includegraphics[width=0.965\textwidth]{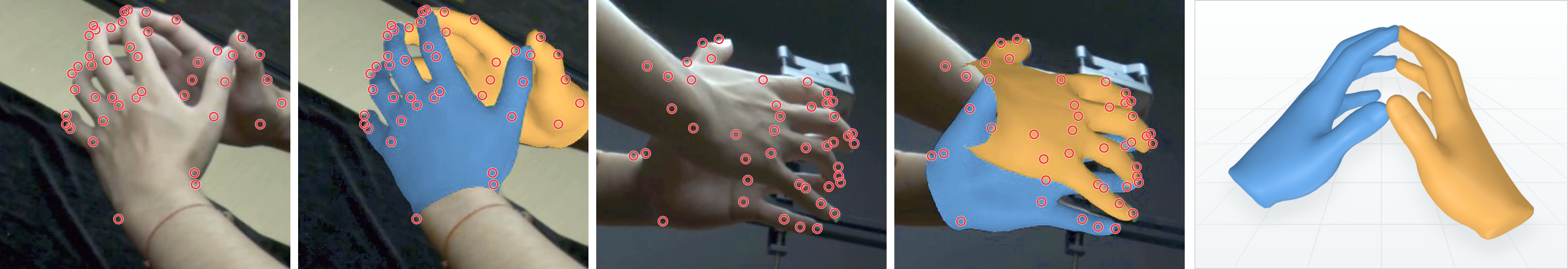}\par\vspace{2pt}
  \leavevmode\vbox to 0.1656\textwidth{\vfil\hbox to 14pt{\hfil\rotatebox{90}{\footnotesize Gestures}\hfil}\vfil}\hspace{2pt}%
  \includegraphics[width=0.965\textwidth]{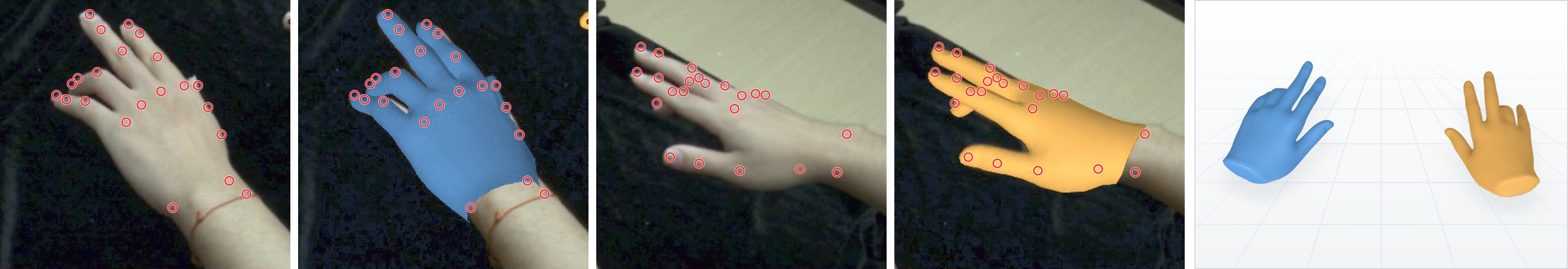}\par\vspace{2pt}
  \leavevmode\vbox to 0.1656\textwidth{\vfil\hbox to 14pt{\hfil\rotatebox{90}{\footnotesize Towel manip.}\hfil}\vfil}\hspace{2pt}%
  \includegraphics[width=0.965\textwidth]{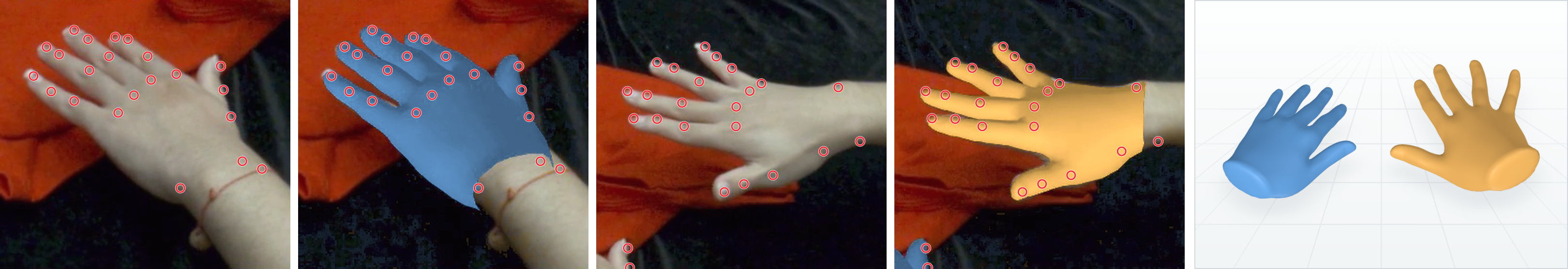}
  \caption{Zero-shot reconstructions on optical motion-capture validation sequences. Projected OptiTrack markers are shown in red on the input crops alongside overlaid MANO meshes from Ours$^*$ and the corresponding 3D reconstructions.}
  \label{fig:suppl_optitrack}
\end{figure*}

\noindent\textbf{Capture rig and synchronization.}
HandQuery-Real is recorded in a custom multi-view tabletop studio (Fig.~\ref{fig:suppl_capture_rig}) integrating $12$ RGB cameras at $1624 \times 1240$ resolution and $30$\,fps with $10$ OptiTrack motion-capture cameras. All cameras are synchronized via the Precision Time Protocol, with an independent LED timing board confirming sub-$50\,\mu\mathrm{s}$ exposure skew across continuous recordings. The full dataset comprises over $2{,}000$ bare-hand manipulation segments totaling roughly $650\mathrm{k}$ multi-view frames.

\noindent\textbf{Evaluation subset.}
From this collection, HandQuery-Eval curates $39$ challenging bimanual segments totaling $16{,}572$ frames across five camera layouts, covering fast and occlusion-heavy interactions such as tossing, catching, tool use, block assembly, handovers, pen spinning, Rubik's cube manipulation, and surface wiping. Whereas reference annotations use all $12$ views to resolve heavy occlusions, benchmark evaluation uses eight- and four-camera subsets to test reconstruction under sparser viewpoints: cameras in each layout are ranked by hand visibility and apparent scale to form the eight-camera subset, from which four views with at least $45^\circ$ angular separation around the workspace are selected for the four-camera setting.

\subsection{Semi-Automatic Annotation Pipeline}
\label{appendix:real_annotation}

Reference MANO annotations for HandQuery-Eval are obtained from all $12$ synchronized RGB views through a four-stage human-in-the-loop workflow. Each subject's hands are first 3D-scanned prior to capture to obtain personalized MANO shape parameters $\boldsymbol{\beta}$. Sapiens2-5B~\cite{khirodkar2026sapiens2} then estimates 2D hand keypoints across all $12$ views for initial multi-view MANO fitting. Human annotators inspect the resulting meshes frame by frame across views, and any flagged frames are refined by an LLM-based agent before returning for manual verification; frames that still fail inspection after iterative refinement are excluded. Leveraging all $12$ surrounding viewpoints during annotation preserves reliable reference geometry even when individual evaluation subsets observe severe hand--object or self-occlusion.

\subsection{Optical Motion-Capture Validation}
\label{appendix:optitrack}

\begin{figure}[t]
  \centering
  \includegraphics[width=\columnwidth]{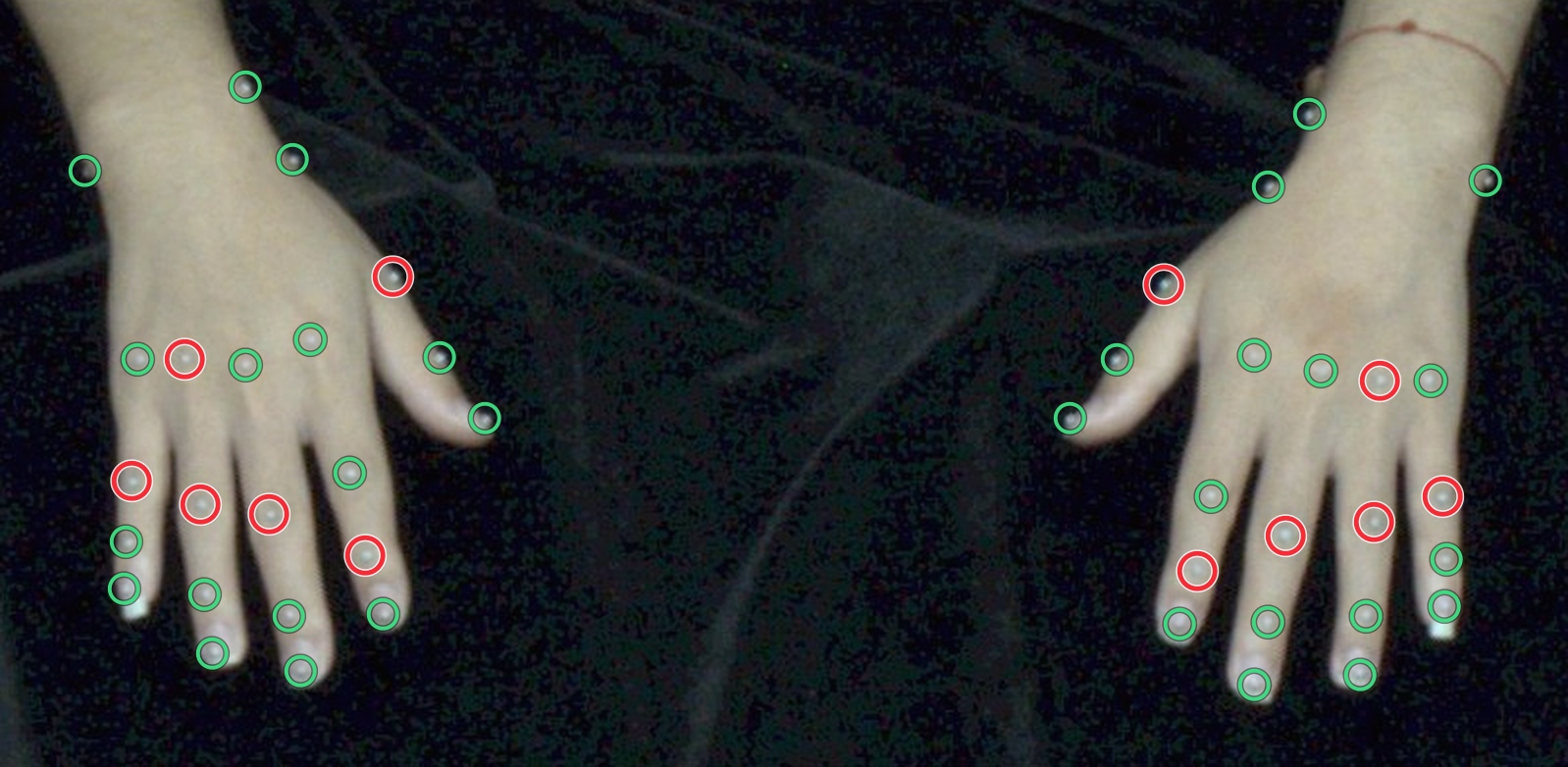}
  \caption{Dorsal optical marker layout with $22$ reflective markers per hand. Sixteen markers in green drive the OptiTrack + MANO reference fit, while six markers in red are held out for evaluation.}
  \label{fig:suppl_marker_layout}
\end{figure}

\begin{table}[t]
  \centering
  \footnotesize
  \setlength{\tabcolsep}{0pt}
  \setlength{\abovecaptionskip}{4pt}
  \setlength{\aboverulesep}{0pt}
  \setlength{\belowrulesep}{0pt}
  \renewcommand{\arraystretch}{1.22}
  \caption{Zero-shot evaluation against held-out OptiTrack markers in world coordinates (mm; best vision result in bold).}
  \label{tab:suppl_optitrack}
  \begin{tabular*}{\columnwidth}{@{\hspace{1pt}}l@{\hspace{6pt}\vrule width \arrayrulewidth\extracolsep{\fill}}ccc@{\hspace{1pt}}}
    \toprule
    Method & Mean$\downarrow$ & Median$\downarrow$ & P90$\downarrow$ \\
    \midrule
    OptiTrack + MANO (16 markers, ref.) & 4.24 & 3.90 & 7.00 \\
    \midrule
    POEM-v2 Large & 7.93 & 7.47 & 11.78 \\
    MAMMA & 7.68 & 7.49 & 11.27 \\
    UA-Fit & 6.81 & 6.65 & 10.39 \\
    WiLoR + Fit & 6.63 & 6.12 & 10.29 \\
    \cmidrule{1-4}
    Ours$^*$ & \textbf{4.83} & \textbf{4.22} & \textbf{7.65} \\
    \bottomrule
  \end{tabular*}
\end{table}

To validate 3D metric accuracy directly against physical measurements, we record nine unseen manipulation actions totaling $8{,}836$ hand-frames with $22$ dorsal reflective markers per hand tracked by the $10$-camera OptiTrack system alongside eight RGB cameras (Fig.~\ref{fig:suppl_marker_layout}). Subject-specific marker attachments are calibrated on separate takes; during evaluation, $16$ markers per hand drive an OptiTrack + MANO reference fit, while six markers spanning all five digits and the dorsum are held out for scoring. All vision methods run zero-shot on the eight RGB cameras using identical multi-view detector boxes without ground-truth crops, and are evaluated directly against the held-out 3D marker positions in calibrated world coordinates without root or Procrustes alignment.

\begin{figure*}[t]
  \centering
  {\small%
  \makebox[16pt]{}%
  \makebox[0.2382\textwidth]{Occluded Target View}\hspace{0.0041\textwidth}%
  \makebox[0.2382\textwidth]{w/o Fusion}\hspace{0.0041\textwidth}%
  \makebox[0.2382\textwidth]{w/ Fusion (Ours)}\hspace{0.0041\textwidth}%
  \makebox[0.2382\textwidth]{Visible Support View}}\par\vspace{2pt}
  \leavevmode\vbox to 0.2382\textwidth{\vfil\hbox to 14pt{\hfil\rotatebox{90}{\footnotesize DexYCB-MV}\hfil}\vfil}\hspace{2pt}%
  \includegraphics[width=0.965\textwidth]{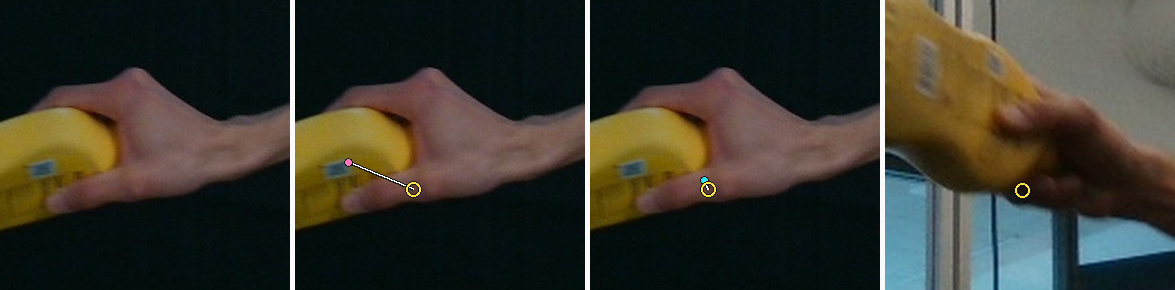}\par\vspace{2.5pt}
  \leavevmode\vbox to 0.2382\textwidth{\vfil\hbox to 14pt{\hfil\rotatebox{90}{\footnotesize ARCTIC-MV}\hfil}\vfil}\hspace{2pt}%
  \includegraphics[width=0.965\textwidth]{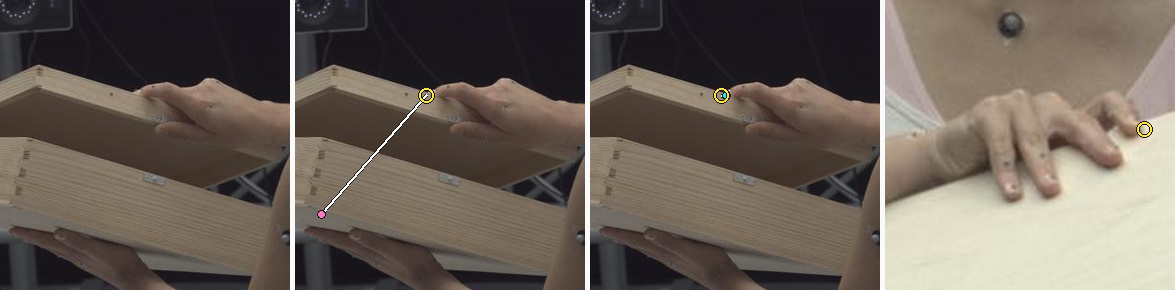}
  \caption{Effect of cross-view fusion on occluded fingertip localization on DexYCB-MV and ARCTIC-MV. Pre-fitting 2D landmark predictions without fusion (pink) and with fusion (cyan) are compared against the projected ground truth (yellow), alongside a synchronized view where the fingertip is visible.}
  \label{fig:suppl_view_fusion}
  \vspace{-5pt}
\end{figure*}

As shown in Table~\ref{tab:suppl_optitrack} and Fig.~\ref{fig:suppl_optitrack}, Ours$^*$ achieves the lowest error among all vision-based methods across all metrics, reducing mean marker error by $27$--$39\%$ compared with all evaluated vision baselines and approaching the $16$-marker optical reference ($4.83$ vs.\ $4.24$\,mm). These results confirm that HandQuery recovers metrically accurate 3D hand motion in unseen capture setups without ground-truth crops or post-hoc alignment.

\section{Efficiency and Multi-View Fusion Analyses}
\label{appendix:more_experiments}

\subsection{Model Size and Inference Cost}
\label{sec:inference_cost}
{\setlength{\belowcaptionskip}{4pt}


\begin{table}[t]
\centering
\caption{Model size and inference cost on eight-view
DexYCB-MV. Timings are measured per synchronized frame.
MANO fitting uses 778 vertices over 72-frame sequences
for WiLoR and HandQuery; UA-Fit uses its own solver,
while POEM-v2 directly predicts meshes.}
\label{tab:inference_cost}

\small
\setlength{\tabcolsep}{2pt}
\begin{tabular*}{\columnwidth}
{@{\extracolsep{\fill}}lrrrr@{}}
\toprule
 & Params & \multicolumn{3}{c}{Time (ms/frame)} \\
\cmidrule(lr){3-5}
Method & (M) & Net & Fit & Total \\
\midrule
POEM-v2 Large & 151.5 & 74.66 & -- & 74.66 \\
WiLoR + Fit   & 640.8 & 74.89 & 11.35 & 86.24 \\
UA-Fit        & 47.2  & 45.97 & 156.91 & 202.88 \\
\midrule
Ours          & 157.9 & 35.44 & 11.35
              & 46.79 \\
\bottomrule
\end{tabular*}
\end{table}}
Table~\ref{tab:inference_cost} compares model size and inference speed on eight-view DexYCB-MV, measured per synchronized frame on a single NVIDIA RTX 4090 after warm-up using ground-truth hand crops (excluding detection and image decoding). At our evaluation stride of $24$ ($T = 48$, half-window overlap), each frame is processed twice in steady state, and Ours takes $35.44$\,ms per frame for the network and $46.79$\,ms including MANO fitting ($1.60\times$ faster than POEM-v2 Large, $1.84\times$ faster than WiLoR + Fit, and $4.34\times$ faster than UA-Fit). With disjoint $48$-frame clips, the same checkpoint takes $17.72$\,ms per frame for the network and $29.07$\,ms including fitting ($\sim 34$\,fps); Ours$^*$ shares the identical architecture and runtime ($35.71$\,ms overlapped, $17.86$\,ms disjoint).

\subsection{Multi-View Fusion Ablation}
\label{appendix:view_fusion}
\begin{table}[ht]
  \centering
  \footnotesize
  \setlength{\tabcolsep}{0pt}
  \setlength{\abovecaptionskip}{4pt}
  \setlength{\aboverulesep}{0pt}
  \setlength{\belowrulesep}{0pt}
  \renewcommand{\arraystretch}{1.22}
  \caption{Cross-view fusion ablation over all $778$ vertices on real-only models, reporting pre-fitting 2D error (px, $\%$) and post-fitting 3D error (mm).}
  \label{tab:suppl_fusion_overall}
  \begin{tabular*}{\columnwidth}{@{\hspace{1pt}}l@{\hspace{4pt}\vrule width \arrayrulewidth\hspace{5pt}}l@{\hspace{5pt}\vrule width \arrayrulewidth\extracolsep{\fill}}cc@{}c@{\hspace{3pt}\vrule width \arrayrulewidth\hspace{3pt}}cc@{\hspace{1pt}}}
    \toprule
    \multicolumn{1}{@{\hspace{1pt}}l@{\hspace{4pt}\hspace{\arrayrulewidth}\hspace{5pt}}}{} & & \multicolumn{2}{c}{2D (pre-fit)} & & \multicolumn{2}{c@{\hspace{1pt}}}{3D (mm)} \\
    \cmidrule{3-4}\cmidrule{6-7}
    Dataset & Variant & EPE$\downarrow$ & PCK@10$\uparrow$ & & MPVPE$\downarrow$ & \multicolumn{1}{c@{\hspace{1pt}}}{PA-V$\downarrow$} \\
    \midrule
    \multirow{2}{*}{DexYCB} & w/o Fusion & 4.11 & 94.0 & & 4.70 & 3.04 \\
     & w/ Fusion & \textbf{3.38} & \textbf{97.4} & & \textbf{4.56} & \textbf{2.91} \\
    \midrule
    \multirow{2}{*}{ARCTIC} & w/o Fusion & 12.26 & 57.8 & & 5.01 & 3.89 \\
     & w/ Fusion & \textbf{9.28} & \textbf{65.4} & & \textbf{4.85} & \textbf{3.78} \\
    \bottomrule
  \end{tabular*}
\end{table}
\noindent\textbf{Setup.}
To isolate the benefit of exchanging multi-view evidence before landmark prediction, we compare our real-only model against an ablated variant trained under identical settings, but with the four cross-view encoder blocks replaced by within-image spatial attention so each view is encoded and decoded independently prior to MANO fitting. Following Sec.~\ref{sec:cams} of the main paper, we evaluate both models on the eight-camera benchmarks DexYCB-MV and ARCTIC-MV.

\begin{table*}[t]
  \centering
  \footnotesize
  \setlength{\tabcolsep}{0pt}
  \setlength{\abovecaptionskip}{4pt}
  \setlength{\aboverulesep}{0pt}
  \setlength{\belowrulesep}{0pt}
  \renewcommand{\arraystretch}{1.20}
  \caption{Temporal consistency comparison on DexYCB-MV, ARCTIC-MV, and eight-camera HandQuery-Eval (best in bold). We report mean $778$-vertex velocity error (Vel, $\mathrm{mm}/\mathrm{f}$) and acceleration error (Acc, $\mathrm{mm}/\mathrm{frame}^2$, and $\mathrm{mm}/\mathrm{s}^2$ at $30$\,fps) without any explicit temporal smoothing in MANO fitting.}
  \label{tab:suppl_temporal}
  \begin{tabular*}{\textwidth}{@{\hspace{1pt}}l@{\hspace{8pt}\vrule width \arrayrulewidth\extracolsep{\fill}}ccc@{}c@{\hspace{5pt}\vrule width \arrayrulewidth\hspace{5pt}}ccc@{}c@{\hspace{5pt}\vrule width \arrayrulewidth\hspace{5pt}}ccc@{\hspace{1pt}}}
    \toprule
     & \multicolumn{3}{c}{DexYCB-MV} & & \multicolumn{3}{c}{ARCTIC-MV} & & \multicolumn{3}{c@{\hspace{1pt}}}{HandQuery-Eval (8 cameras)} \\
    \cmidrule{2-4}\cmidrule{6-8}\cmidrule{10-12}
    Method & Vel$\downarrow$ & Acc$\downarrow$ & mm/s$^2\downarrow$ & & Vel$\downarrow$ & Acc$\downarrow$ & mm/s$^2\downarrow$ & & Vel$\downarrow$ & Acc$\downarrow$ & \multicolumn{1}{c@{\hspace{1pt}}}{mm/s$^2\downarrow$} \\
    \midrule
    POEM-v2 & 3.06 & 4.74 & 4269 & & 1.34 & 1.69 & 1517 & & 2.13 & 2.92 & 2632 \\
    UA-Fit & 3.01 & 4.62 & 4155 & & 1.42 & 1.76 & 1588 & & 1.64 & 2.23 & 2010 \\
    WiLoR + Fit & 2.62 & 3.92 & 3532 & & 1.80 & 2.51 & 2257 & & 1.20 & 1.69 & 1524 \\
    MAMMA & 2.18 & 3.21 & 2887 & & 0.99 & 1.20 & 1076 & & 0.96 & 1.32 & 1189 \\
    \cmidrule{1-12}
    Ours ($T = 1$, no temp.\ attn.) & 2.65 & 4.12 & 3710 & & 0.95 & 1.16 & 1042 & & 1.88 & 2.73 & 2458 \\
    Ours ($T = 48$, real-only) & 2.01 & 2.91 & 2618 & & 0.91 & 1.02 & 922 & & 1.10 & 1.43 & 1285 \\
    Ours$^*$ ($T = 48$, real + syn.) & \textbf{1.95} & \textbf{2.81} & \textbf{2530} & & \textbf{0.82} & \textbf{0.90} & \textbf{806} & & \textbf{0.74} & \textbf{0.97} & \textbf{876} \\
    \bottomrule
  \end{tabular*}
  \vspace{-7pt}
\end{table*}

\noindent\textbf{Overall 2D and 3D accuracy.}
As shown in Table~\ref{tab:suppl_fusion_overall}, cross-view fusion reduces pre-fitting 2D endpoint error across all vertices by $17.9\%$ on DexYCB-MV and $24.3\%$ on ARCTIC-MV, and consistently lowers post-fitting 3D error on both benchmarks. The relative gain after fitting is more moderate under full eight-camera coverage because uncertainty weighting, visibility gating, and robust Geman--McClure optimization already suppress isolated per-view outliers when sufficient unobstructed views remain.

\begin{table}[ht]
  \centering
  \footnotesize
  \setlength{\tabcolsep}{0pt}
  \setlength{\abovecaptionskip}{4pt}
  \setlength{\aboverulesep}{0pt}
  \setlength{\belowrulesep}{0pt}
  \renewcommand{\arraystretch}{1.22}
  \caption{Visibility-stratified 2D fingertip localization accuracy without and with cross-view fusion.}
  \label{tab:suppl_fusion_fingertip}
  \begin{tabular*}{\columnwidth}{@{\hspace{1pt}}c@{\hspace{3pt}\vrule width \arrayrulewidth\hspace{4pt}}l@{\hspace{5pt}\vrule width \arrayrulewidth\extracolsep{\fill}}ccc@{\hspace{1pt}}}
    \toprule
    \multicolumn{1}{@{\hspace{1pt}}c@{\hspace{3pt}\hspace{\arrayrulewidth}\hspace{4pt}}}{} & Subset & EPE (px)$\downarrow$ & $\Delta$ & PCK@10$\uparrow$ \\
    \midrule
    \multirow{4}{*}{\rotatebox[origin=c]{90}{DexYCB}} & Visible & 3.66 $\rightarrow$ \textbf{3.23} & $-12.0\%$ & 95.3 $\rightarrow$ \textbf{97.1} \\
     & Occluded (all) & 6.30 $\rightarrow$ \textbf{4.56} & $-27.5\%$ & 82.1 $\rightarrow$ \textbf{92.0} \\
     & \;\;other-view vis. & 6.10 $\rightarrow$ \textbf{4.29} & $-29.6\%$ & 83.3 $\rightarrow$ \textbf{93.5} \\
     & \;\;all-view hidden & 7.29 $\rightarrow$ \textbf{5.93} & $-18.6\%$ & 76.1 $\rightarrow$ \textbf{84.6} \\
    \midrule
    \multirow{4}{*}{\rotatebox[origin=c]{90}{ARCTIC}} & Visible & 8.08 $\rightarrow$ \textbf{6.66} & $-17.6\%$ & 81.1 $\rightarrow$ \textbf{83.7} \\
     & Occluded (all) & 20.19 $\rightarrow$ \textbf{11.35} & $-43.8\%$ & 41.5 $\rightarrow$ \textbf{56.2} \\
     & \;\;other-view vis. & 20.08 $\rightarrow$ \textbf{11.24} & $-44.1\%$ & 41.7 $\rightarrow$ \textbf{56.6} \\
     & \;\;all-view hidden & 21.89 $\rightarrow$ \textbf{13.04} & $-40.4\%$ & 38.8 $\rightarrow$ \textbf{50.2} \\
    \bottomrule
  \end{tabular*}
\end{table}

\noindent\textbf{Occluded-fingertip disambiguation.}
Stratifying fingertip predictions by visibility in Table~\ref{tab:suppl_fusion_fingertip} shows that these gains stem primarily from occluded fingertips (70.1\% and 55.5\% of observations on DexYCB-MV and ARCTIC-MV), which account for approximately 90\% of the total fingertip error reduction on both datasets. Whereas visible fingertips improve moderately, error on occluded fingertips drops by $27.5\%$ on DexYCB-MV and $43.8\%$ on ARCTIC-MV, with the largest reductions occurring when a fingertip occluded in the target view is visible in at least one other camera. As illustrated in Fig.~\ref{fig:suppl_view_fusion}, independent per-view predictions often drift onto occluding objects or the background, whereas cross-view fusion leverages unobstructed views to localize occluded fingertips accurately before model fitting.

\section{Temporal Consistency Analysis}
\label{appendix:temporal}

\noindent\textbf{Metrics and evaluation protocol.}
To evaluate temporal stability alongside spatial accuracy, we measure per-vertex velocity and acceleration errors in calibrated world coordinates over all consecutive frame pairs and triplets of the evaluated sequences. With $\Delta \mathbf{X}_t = \mathbf{X}_t - \mathbf{X}_{t-1}$ and $\Delta^2 \mathbf{X}_t = \mathbf{X}_{t+1} - 2\mathbf{X}_t + \mathbf{X}_{t-1}$, we compute
{\setlength{\abovedisplayskip}{6pt}\setlength{\belowdisplayskip}{6pt}%
\begin{align}
  e^{\mathrm{vel}}_t &= \|\Delta\hat{\mathbf{X}}_t - \Delta\mathbf{X}^{\mathrm{gt}}_t\|_2,
  \label{eq:suppl_vel} \\
  e^{\mathrm{acc}}_t &= \|\Delta^2\hat{\mathbf{X}}_t - \Delta^2\mathbf{X}^{\mathrm{gt}}_t\|_2,
  \label{eq:suppl_acc}
\end{align}}%
averaged over all $778$ MANO vertices. Measuring error relative to ground-truth derivatives rather than raw smoothness penalizes jitter without rewarding over-smoothing of rapid hand motion. Because our MANO fitter uses no explicit temporal regularization, temporal consistency reflects the learned spatiotemporal representation and uncertainty-weighted multi-view predictions.

\noindent\textbf{Comparison with baselines and single-frame ablation.}
As shown in Table~\ref{tab:suppl_temporal}, Ours$^*$ achieves the lowest velocity and acceleration errors across DexYCB-MV, ARCTIC-MV, and zero-shot HandQuery-Eval, reducing acceleration error by $12$--$27\%$ over MAMMA, the strongest frame-wise baseline, and by $28$--$67\%$ over the remaining methods. Comparing the real-only model at $T = 48$ against its single-frame variant at $T = 1$ isolates the contribution of temporal attention, lowering acceleration error by $11.5$--$47.7\%$ across the three benchmarks under identical training data and fitting. Furthermore, the predicted acceleration magnitudes of Ours$^*$ match ground-truth motion statistics rather than damping rapid movements, whereas frame-wise baselines exhibit substantially higher temporal jitter; see the supplementary webpage for multi-view video comparisons.

\begin{table}[ht]
  \centering
  \footnotesize
  \setlength{\tabcolsep}{0pt}
  \setlength{\abovecaptionskip}{4pt}
  \setlength{\aboverulesep}{0pt}
  \setlength{\belowrulesep}{0pt}
  \renewcommand{\arraystretch}{1.22}
  \caption{Sliding-window boundary continuity for Ours$^*$ on sequences exceeding $48$ frames, comparing velocity and acceleration errors at window seams against interior frames and MAMMA's interior acceleration error.}
  \label{tab:suppl_seam}
  \begin{tabular*}{\columnwidth}{@{\hspace{1pt}}l@{\hspace{4pt}\vrule width \arrayrulewidth\extracolsep{\fill}}cc@{}c@{\hspace{3pt}\vrule width \arrayrulewidth\hspace{3pt}}c@{\hspace{1pt}}}
    \toprule
     & \multicolumn{2}{c}{Ours$^*$ (Seam / Intr.)} & & \multicolumn{1}{c@{\hspace{1pt}}}{MAMMA} \\
    \cmidrule{2-3}\cmidrule{5-5}
    Dataset & Vel$\downarrow$ & Acc$\downarrow$ & & \multicolumn{1}{c@{\hspace{1pt}}}{Intr.\ Acc$\downarrow$} \\
    \midrule
    DexYCB-MV & 2.12 / 1.94 {\scriptsize\textcolor{black!55}{+0.18}} & 3.13 / 2.78 {\scriptsize\textcolor{black!55}{+0.35}} & & 3.19 \\
    ARCTIC-MV & 0.97 / 0.82 {\scriptsize\textcolor{black!55}{+0.15}} & 1.04 / 0.88 {\scriptsize\textcolor{black!55}{+0.16}} & & 1.19 \\
    HandQuery-Eval & 0.89 / 0.73 {\scriptsize\textcolor{black!55}{+0.16}} & 1.10 / 0.96 {\scriptsize\textcolor{black!55}{+0.14}} & & 1.32 \\
    \bottomrule
  \end{tabular*}
  \vspace{-8pt}
\end{table}

\begin{figure*}[t]
  \centering
  {\small%
  \makebox[16pt]{}%
  \makebox[0.1890\textwidth]{Input}\hspace{0.0046\textwidth}%
  \makebox[0.1890\textwidth]{Ours$^*$}\hspace{0.0065\textwidth}%
  \makebox[0.1890\textwidth]{3D Human (Ours$^*$)}\hspace{0.0046\textwidth}%
  \makebox[0.1890\textwidth]{Sharpa (UMR)}\hspace{0.0046\textwidth}%
  \makebox[0.1890\textwidth]{Wuji (UMR)}}\par\vspace{2pt}
  \leavevmode\vbox to 0.1434\textwidth{\vfil\hbox to 14pt{\hfil\rotatebox{90}{\footnotesize Pen spinning}\hfil}\vfil}\hspace{2pt}%
  \includegraphics[width=0.965\textwidth]{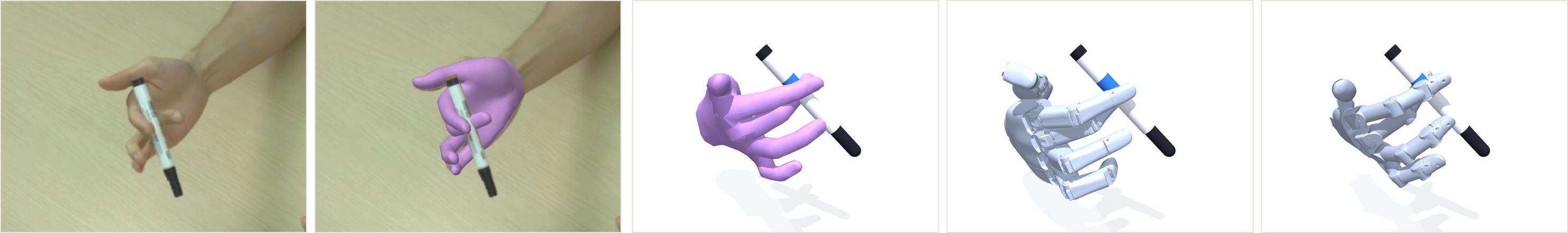}\par\vspace{2.5pt}
  \leavevmode\vbox to 0.1434\textwidth{\vfil\hbox to 14pt{\hfil\rotatebox{90}{\footnotesize Rubik's cube}\hfil}\vfil}\hspace{2pt}%
  \includegraphics[width=0.965\textwidth]{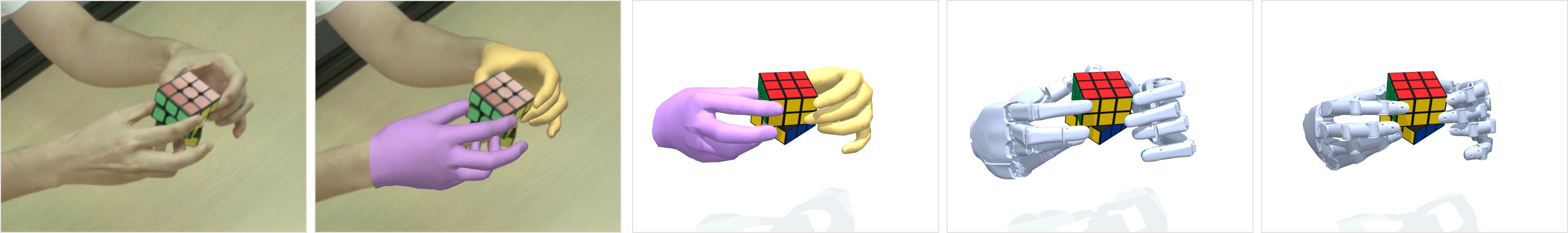}\par\vspace{2.5pt}
  \leavevmode\vbox to 0.1434\textwidth{\vfil\hbox to 14pt{\hfil\rotatebox{90}{\footnotesize Apple manip.}\hfil}\vfil}\hspace{2pt}%
  \includegraphics[width=0.965\textwidth]{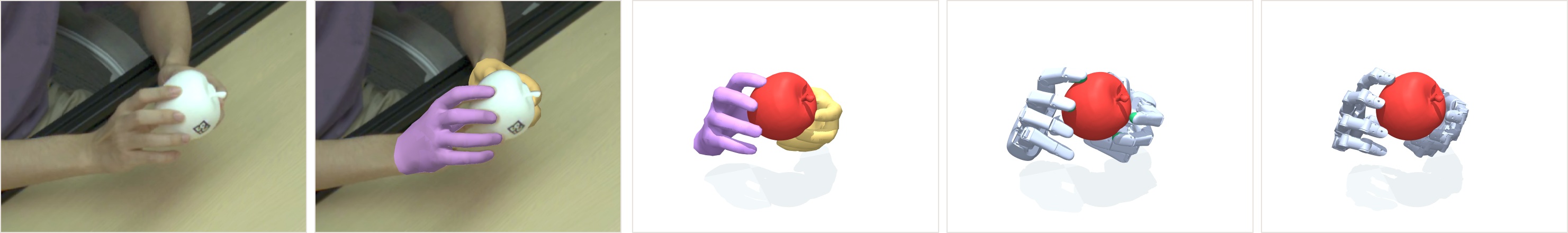}
  \caption{Downstream dexterous robot hand retargeting from bare-hand multi-view videos. From zero-shot 4D MANO trajectories reconstructed by Ours$^*$ (left: yellow, right: violet), we apply UMR~\cite{cao2026umr} to retarget fine in-hand and bimanual manipulation motions onto two distinct dexterous robot hands (\textit{Sharpa} and \textit{Wuji}).}
  \label{fig:suppl_retarget}
  \vspace{-6pt}
\end{figure*}

\noindent\textbf{Sliding-window boundary continuity.}
For sequences exceeding $48$ frames, overlapping windows (stride $24$) are fused by inverse-variance weighting prior to fitting. Table~\ref{tab:suppl_seam} compares window-transition seams, where the covering window set changes, against interior frames. Without boundary tapering, transitions add only a sub-millimeter step of $0.15$--$0.18\,\mathrm{mm}/\mathrm{f}$ in velocity error and $0.14$--$0.35\,\mathrm{mm}/\mathrm{f}^2$ in acceleration error, and seam acceleration error for Ours$^*$ still remains below the interior average of every frame-wise baseline, including MAMMA.

\section{Additional Qualitative Comparisons}
\label{appendix:qualitative}

\begin{figure*}[p]
  \centering
  {\small%
  \makebox[0.1428\textwidth]{Image}%
  \makebox[0.1428\textwidth]{GT}%
  \makebox[0.1428\textwidth]{WiLoR + Fit}%
  \makebox[0.1428\textwidth]{POEM-v2}%
  \makebox[0.1428\textwidth]{UA-Fit}%
  \makebox[0.1428\textwidth]{MAMMA}%
  \makebox[0.1428\textwidth]{Ours$^*$}}\par\vspace{2pt}
  \includegraphics[width=\textwidth]{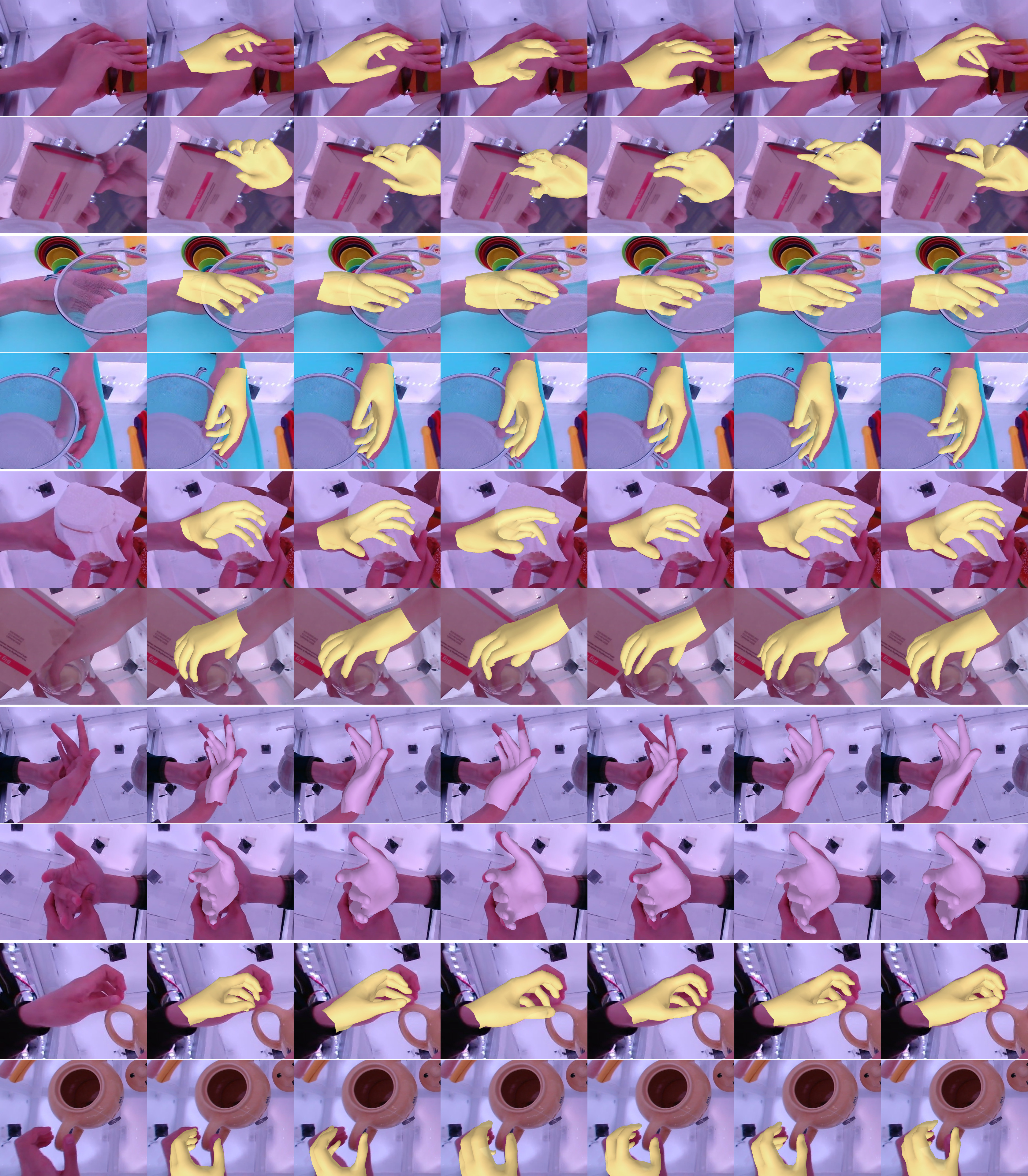}
  \caption{Additional zero-shot results on GigaHands~\cite{fu2025gigahands}. Five unseen examples are shown, each from two synchronized views. Left hands are shown in yellow and right hands in violet.}
  \label{fig:suppl_qual_zs}
\end{figure*}

\begin{figure*}[p]
  \centering
  {\small%
  \makebox[0.1428\textwidth]{Image}%
  \makebox[0.1428\textwidth]{GT}%
  \makebox[0.1428\textwidth]{WiLoR + Fit}%
  \makebox[0.1428\textwidth]{POEM-v2}%
  \makebox[0.1428\textwidth]{UA-Fit}%
  \makebox[0.1428\textwidth]{MAMMA}%
  \makebox[0.1428\textwidth]{Ours$^*$}}\par\vspace{2pt}
  \includegraphics[width=\textwidth]{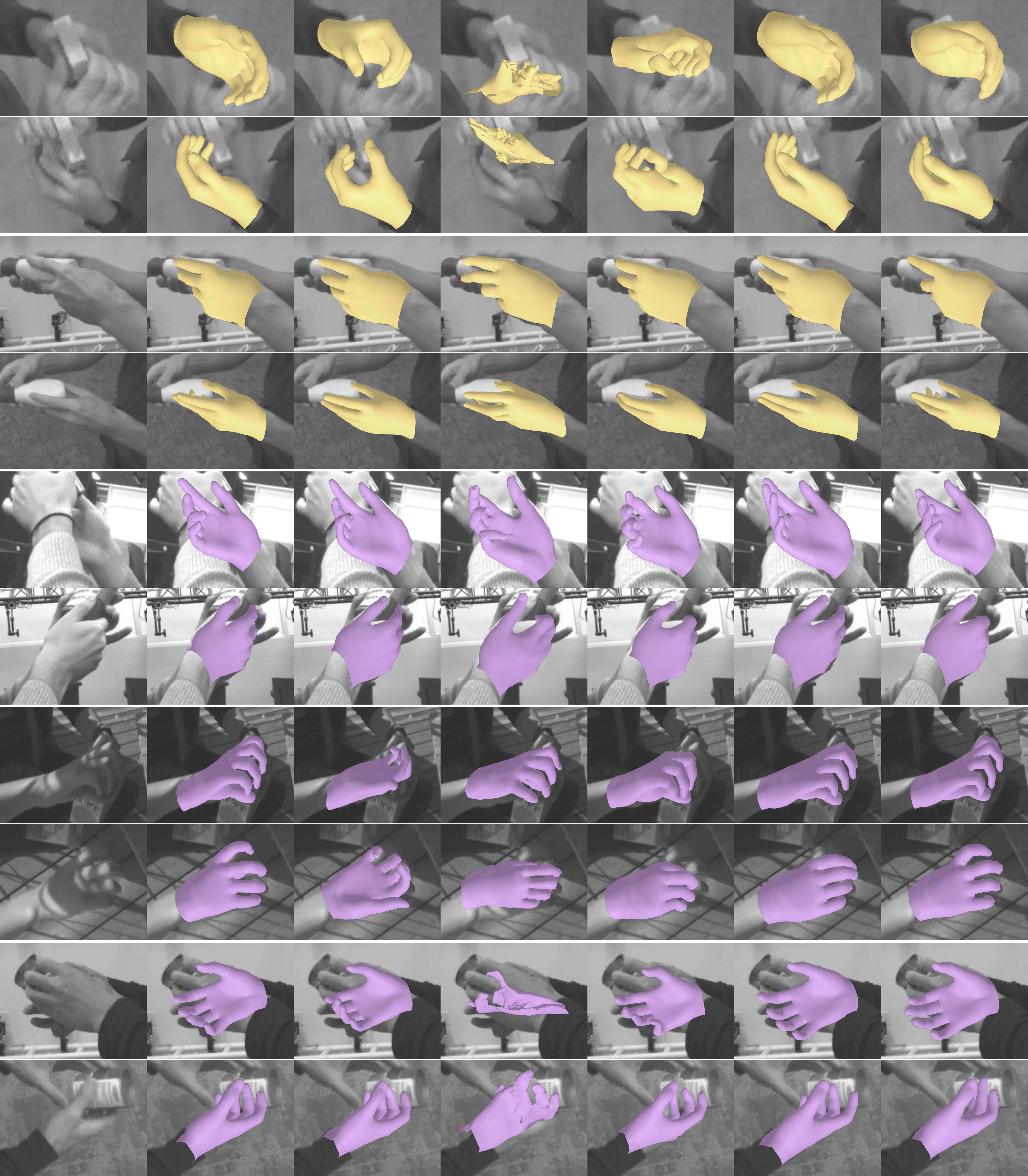}
  \caption{Additional zero-shot results on SHOW3D~\cite{rim2026show3d}. Five unseen examples are shown, each from two synchronized views. Left hands are shown in yellow and right hands in violet.}
  \label{fig:suppl_qual_zs_show3d}
\end{figure*}

Figs.~\ref{fig:suppl_qual_zs} and~\ref{fig:suppl_qual_zs_show3d} provide additional zero-shot comparisons on GigaHands~\cite{fu2025gigahands} and SHOW3D~\cite{rim2026show3d}, each showing five unseen sequences from two synchronized views. Under fine finger articulation, bimanual contact, and severe object occlusion, frame-wise and monocular-plus-fitting baselines often collapse occluded digits, interpenetrate fingers, or drift across views, whereas Ours$^*$ recovers view-consistent 3D hand geometry that often aligns closer to the images than the reference annotations; synchronized zero-shot videos are provided on the supplementary webpage.


\section{Dexterous Robot Hand Retargeting}
\label{appendix:retargeting}

Metrically accurate and temporally coherent bare-hand 4D reconstruction provides a useful foundation for markerless human-to-robot motion retargeting. To demonstrate this downstream capability, we retarget zero-shot 4D MANO trajectories reconstructed by Ours$^*$ on unseen studio captures onto two morphologically distinct dexterous robot-hand embodiments, \textit{Sharpa} and \textit{Wuji}, using Unified Motion Retargeting (UMR)~\cite{cao2026umr}. Rather than relying on hand-crafted joint-pair mappings across mismatched kinematic topologies, UMR learns dense point-cloud correspondences between human and robot hand surfaces and solves constrained surface-and-contact optimization to transfer fine interaction geometry. As shown in Fig.~\ref{fig:suppl_retarget}, across dynamic in-hand pen spinning, bimanual Rubik's cube rotation, and object manipulation, the dense surface fidelity and low temporal jitter of Ours$^*$ preserve finger--object contact structure and yield coherent trajectories across both robot embodiments; synchronized multi-view videos and interactive 3D visualizations are provided on the supplementary webpage.



\section{Ethics Statement}
\label{appendix:ethics}

All recordings in HandQuery-Real, HandQuery-Eval, and the optical motion-capture validation sequences involve only non-invasive, everyday tabletop hand--object interactions posing minimal risk to participants. Prior to 3D hand scanning and multi-view recording, all adult participants were informed of the capture procedures and intended research use, and provided informed consent for academic publication and dataset release. Formal institutional ethics review was not sought, as the study was considered by the authors to involve minimal risk and voluntary participation. No directly identifying metadata are included in the released datasets, and any incidental facial regions appearing in wide-angle views are anonymized prior to visualization and public release.

{
    \small
    \bibliographystyle{ieeenat_fullname}
    \bibliography{main}
}  


\end{document}